\documentclass[journal]{IEEEtran}

\usepackage{makecell}
\usepackage{multirow}
\usepackage[english]{babel}
\newcommand\mgape[1]{\gape{$\vcenter{\hbox{#1}}$}}
\newcommand{\tabincell}[2]{\begin{tabular}{@{}#1@{}}#2\end{tabular}}

\usepackage{graphicx}
\usepackage[namelimits]{amsmath} 
\usepackage{amssymb}             
\usepackage{amsfonts}            
\usepackage{mathrsfs}            
\usepackage{bm}            
\usepackage{subfigure}
\usepackage{caption}
\usepackage{multirow}
\usepackage{bbding}
\usepackage{cite}
\usepackage{hyperref}

\emergencystretch=\maxdimen
\begin{document}

\title{To See in the Dark: N2DGAN for Background Modeling in Nighttime Scene}
\author{Zhenfeng Zhu,
        Yingying Meng,
        Deqiang Kong,
        Xingxing Zhang,

       Yandong Guo, and Yao Zhao, ~\IEEEmembership{Senior Member,~IEEE}
\thanks{This work was supported in part by Science and Technology Innovation 2030 - "New Generation Artificial Intelligence" Major Project under
Grant 2018AAA0102101, in part by the National Natural Science Foundation of China under Grant 61976018 and Grant 61532005, and in part by the Fundamental Research Funds for the Central Universities under Grant 2018JBZ001}
\thanks{Zhenfeng Zhu,
        Yingying Meng
        Xingxing Zhang,
        and~Yao Zhao are with the Institute of Information Science, Beijing Jiaotong University, Beijing, 100044, China, and also with Beijing Key Laboratory of Advanced Information Science and Network Technology, Beijing,
100044, China. E-mail: \{zhfzhu, mengyingying, zhangxing, yzhao\}@bjtu.edu.cn.}
\thanks{Deqiang Kong is with Microsoft Multimedia, Beijing, China. Email: kodeqian@microsoft.com.}
\thanks{Yandong Guo is with OPPO Research Institute, Beijing, China. Email: guoyandong@oppo.com}
\thanks{Manuscript received April 19, 2015; revised August 26, 2015.}}

\markboth{Journal of \LaTeX\ Class Files,~Vol.~14, No.~8, August~2015}%
{Shell \MakeLowercase{\textit{et al.}}: Bare Demo of IEEEtran.cls for IEEE Journals}


\maketitle

\begin{abstract}
Due to the deteriorated conditions of \mbox{illumination} lack and uneven lighting, the performance of traditional background modeling methods is greatly limited for the surveillance of nighttime video.
  To make background modeling under nighttime scene performs as well as in daytime condition, we put forward a promising generation-based background modeling framework for foreground surveillance.
  With a pre-specified daytime reference image as background frame, the {\bfseries GAN} based generation model, called {\bfseries N2DGAN}, is trained to transfer each frame of {\bfseries n}ighttime video {\bfseries to} a virtual {\bfseries d}aytime image with the same scene to the reference image except for the foreground part.
  Specifically, to balance the preservation of background scene and the foreground object(s) in generating the virtual daytime image, we presented a two-pathway generation model, in which the global and local sub-networks were well combined with spatial and temporal consistency constraints.
  For the sequence of generated virtual daytime images, a multi-scale Bayes model was further proposed to characterize pertinently the temporal variation of background.
  We manually labeled ground truth on the collected nightime video datasets for performance evaluation. The impressive results illustrated in both the main paper and supplementary show the effectiveness of our proposed approach.
\end{abstract}


\IEEEpeerreviewmaketitle

\section{Introduction}
Background modeling originates in numerous applications, especially in visual surveillance \cite{Barnich2011ViBe,Elgammal2002Background,Kim2005Real,Zivkovic2004Improved,DBLP:journals/corr/abs-1801-02225, DBLP:journals/tits/YangLLZCL18, DBLP:journals/access/ZengZ18}.
\begin{figure}[!ht]
\centering
\includegraphics[width=0.5\textwidth]{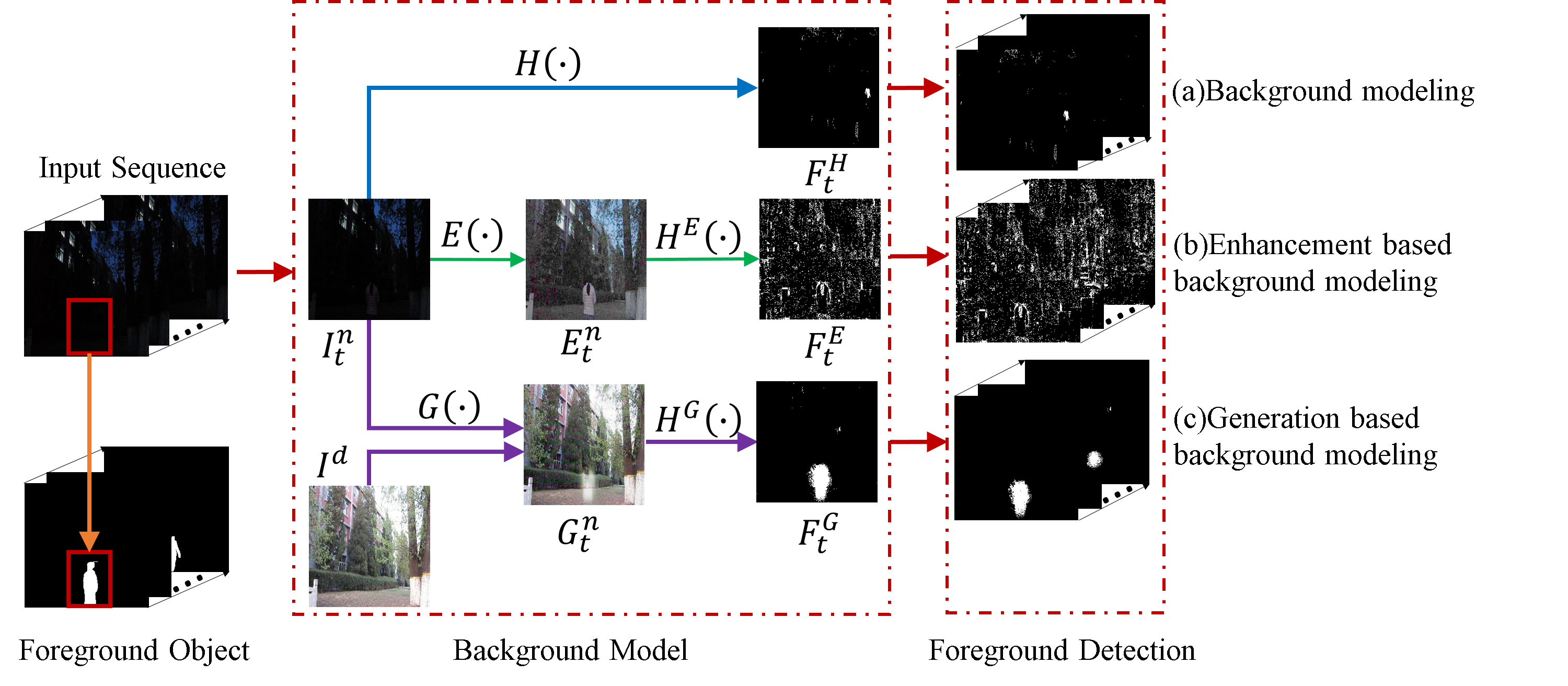}
\caption{Flowchart of three kinds of background modeling methods for foreground object detection in nighttime surveillance video. (a) Conventional . (b) Enhancement-based. (c) Generation-based.}
\label{fig_overview}
\end{figure}
In the last decades, state-of-the-art approaches for background modeling have been proposed for visual surveillance under daytime scenes.
On a whole, they are popularly dominated by a family of statistical based methods, like GMM \cite{Zivkovic2004Improved} and KDE \cite{Elgammal2002Background}.
Besides, Codebook \cite{Kim2005Real} and ViBe \cite{Barnich2011ViBe} are also two representative methods that achieve good performance for modeling the background. Most recently, several deep learning based works \cite{DBLP:journals/corr/abs-1801-02225, DBLP:journals/tits/YangLLZCL18, DBLP:journals/access/ZengZ18} for background modeling were also proposed.

Despite the achievements by these approaches, all of them face quite a challenge in the case of illumination lack and uneven lighting at night, especially in the presence of dynamic background, change of light, and some extreme weather conditions such as rain, snow and fog.
As we can see from Fig.\ref{fig_overview} (a), the conventional background modeling methods, like GMM, etc., fail to distinguish the foreground object from background due to the deteriorated condition of illumination lack. To deal with such a case, an intuitive way as shown in Fig.\ref{fig_overview} (b) is to perform image enhancement through $E(\cdot)$ first, and then build background model $H^{E}(\cdot)$ on the bases of the enhanced frames $E^n_t$'s, just like background modeling under daytime scene. However, since these enhancement methods \cite{Jobson2002A, Ketcham1976Real, Yan2016Nighttime} are not task-driven, they usually lose sight of pixel-wise consistency of inter-frame, and yet it is of great significance for background modeling.

Although the captured images can be brightened by equipping a light with the camera, there are still deficiencies in the use of these devices. First, it will inevitably increase the cost of surveillance. Second, compared with the imaging under the natural light in the daytime, the visual quality by this way is still unsatisfactory. For example, some problems including serious noise, blurring, and unbalanced illumination will further bring difficulties for some vision processing tasks including object detection. In addition, it also won't work well for monitoring distant object.

To address this issue, we make a novel contribution in integrating generative model into background modeling. Fig.\ref{fig_overview} (c) shows the proposed generation-based background modeling framework. With a pre-specified daytime reference image $ I^d$ as ground-truth background frame, the generation model $G(\cdot)$ is trained for transferring each frame $I_{t}^{n}$ of nighttime video to a virtual daytime image $G_{t}^{n}$ with the same scene to the reference image except for the foreground region. Furthermore, the background model $H_{G}(\cdot)$ can be built to obtain $F_{t}^{G}$ with the detected foreground object. In fact, the unique reference image plays a significant role for enforcing the pixel-wise temporal consistency of inter-frames in the generation of virtual daytime images.

To the best of our knowledge, this paper is one of the first attempts to introduce GANs based deep learning network for background modeling. In summary, the following points highlight several contributions of the paper:

\begin{itemize}
\item{
  This paper proposes a reasonable and innovative solution, i.e., N2DGAN, to the longstanding problem of foreground object detection under nighttime scene. As a promising generation-based framework, it makes background modeling work as well as in daytime condition.
 }
\item{
To simultaneously preserve background scene and the foreground object(s) in generating the virtual daytime image, we present a two-pathway generation model, in which the global and local sub-networks are seamlessly combined with spatial and temporal consistency constraints.}
\item{
    For the sequence of generated virtual daytime images, a multi-scale Bayes model is proposed to characterize pertinently the temporal variation of background. Thus, while suppressing effectively noise coming from virtual daytime image generation, we can ensure the favorable detection of foreground objects.}
\item{
  We collect a benchmark dataset including indoor and outdoor scenes with manually labeled ground truth, which can serve as a good benchmark for the research community.
  }
\end{itemize}

\section{Related Work}
{\bfseries Nighttime Image Enhancement.}
Here we simply divide image enhancement methods into two categories: reference based and non-reference based methods.

Non-reference based methods mainly focus on how to improve low contrast images.
As a naive method, Histogram Equalization (HE)\cite{Ketcham1976Real} spreads out the most frequent intensity values, thus gaining a higher contrast for the areas of lower contrast. The purpose of Retinex based image enhancement [MSR]\cite{Jobson2002A} is to estimate the illumination from original image, thereby decomposing reflectance image and eliminating the influence of uneven illumination. In recent years, some deep learning based low light image enhancement approaches were also proposed, such as LIME \cite{DBLP:Guo2016}, LLNet \cite{DBLP:Lore2017}, and Struct \cite{DBLP:Li2018}. Although they generally can achieve better perceptual quality than HE and MSR, they highly depend on the amount of training data.

Reference based methods \cite{Cai2006Context, Liang2012Image, Raskar2004Image} usually combine images of a scene at different time intervals by image fusion.
These methods usually produce unnatural effects in the enhanced images.
Besides, it would increase signal to noise ratio, which is adverse for further video analysis and applications such as foreground detection.

{\bfseries Generative Adversarial Nets.}
As a novel way to train generative models, GANs \cite{goodfellow2014generative} proposed by Goodfellow et al. has received extensive applications in various of visual tasks\cite{iizuka2017globally,yeh2016semantic,taigman2016unsupervised,Pixel2Pixel2017,ledig2017photo,Zheng2019}.
In \cite{iizuka2017globally}, GANs was applied for image completion with globally and locally consistent adversarial training. \cite{yeh2016semantic} used back-propagation on a pretrained image generative network for image inpainting. To transfer the original image into a cartoon style, domain transfer network (DTN) \cite{taigman2016unsupervised} was proposed. In \cite{ledig2017photo,Zheng2019}, GANs has been employed for image super-resolution and image deblurring. Recently, a general-purpose solution to image-to-image translation based on conditional adversarial networks, also known as pixel2pixel network, was proposed in \cite{Pixel2Pixel2017}, and shows good performances on a variety of tasks like photo generation and semantic segmentation. In our previous work\cite{meng2019}, a generative adversarial networks (GANs) based framework for nighttime image enhancement was proposed.

{\bfseries Background Modeling Algorithms.}
Broadly speaking, background modeling methods can be divided into two categories: pixel-based methods and block-based methods.

One of the most popular pixel-based methods is Gaussian mixture models(GMM) \cite{Kaewtrakulpong2002An, Stauffer1999Adaptive, Zivkovic2004Improved}. It models the distribution at each pixel observed over time using a summation of weighted Gaussian distribution.
Such methods generally perform well with the multi-modal nature of many practical situations. However, if high or low frequency changes appear in the background, the model can't be adaptively tuned in time and even may miss some information about fast moving objects.
Consequently, Elgammal et al.\cite{Elgammal2000Non} have developed a non-parametric background model, which estimates the probability of observing pixel intensity values based on a sample of intensity values for each pixel.

Different from GMM and KDE, some deep learning based works \cite{DBLP:journals/corr/abs-1801-02225, DBLP:journals/tits/YangLLZCL18, DBLP:journals/access/ZengZ18} for background modeling have also been proposed in recent years. But these models are all supervised and require many manually labeled data for model training. Thus, they obviously lack of scalability to a scene unknown beforehand, which also means their performances greatly depend on the collected training datasets.


Block-based methods \cite{heikkila2006texture, DBLP:conf/cvpr/LiaoZKPL10} divide each frame into multiple overlapped or non-overlapped small blocks, and then model the background using the features of each block. Compared with pixel-based methods, the image blocks can capture more spatial distribution information, which makes block-based methods insensitive to the local shift in the background. However, the detection performance will largely depend on the block-dividing technique, especially for small moving targets.
\begin{figure*}[!ht]
\centering
\includegraphics[width=0.8\textwidth]{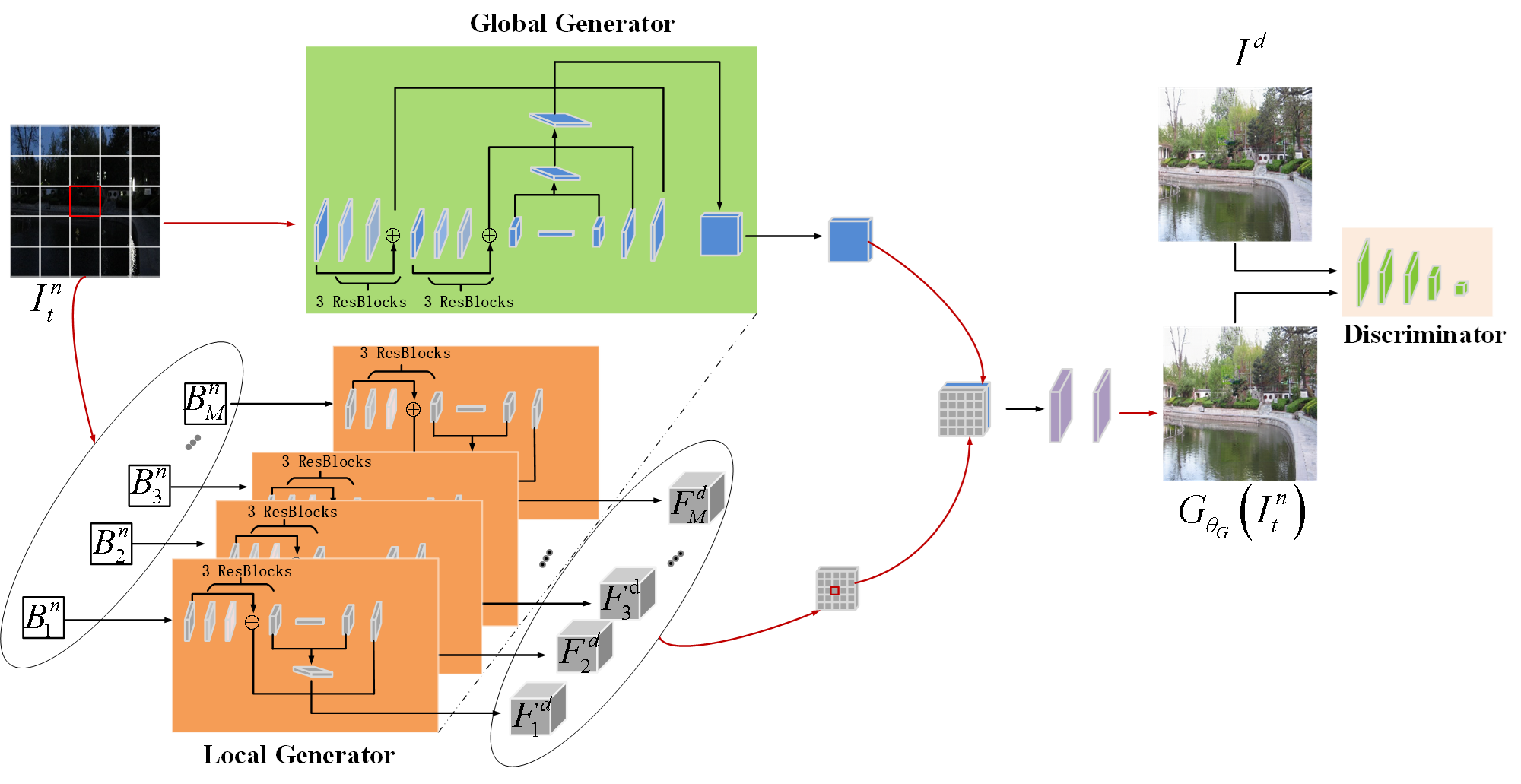}
\caption{The network architecture of N2DGAN. $I^{n}$ is the input nighttime image, and we divide it into $M$ blocks ($B_{i}^{n}, i=1,2,...,M$) as the input of each local generator. Then, the output of global generator and local generators are concatenated together. Finally, after two convolutional layers, the output is the daytime image $G_{\theta_{G}}(I^{n})$, where $I^{d}$ is the reference daytime image. More details about our model architecture are provided in the appendix.}
\label{fig_model1}
\end{figure*}
\section{Nighttime to Daytime Generative Adversarial Networks (N2DGAN)}
To maintain spatial and temporal consistency in the generation process,
our goal is to train a generation model $G(\cdot)$ as in Fig.\ref{fig_overview} to transfer each frame $I_{t}^{n}$ of nighttime video to a virtual daytime image with the same scene to the unique reference image $I^{d}$ except for the foreground region.
Specifically, this generation problem can be formulated as:
\begin{equation}
\hat{\theta}_G = \mathop{argmin}_{\theta_G} \frac{1}{N}\sum_{t=1}^{N} L(G_{\theta_G}(I^{n}_{t}), I^{d})
\label{e1}
\end{equation}
where $N$ is the number of training pairs, $L(\cdot,\cdot)$ denotes a weighted combination of several loss components.

For the intent of learning the generation function $G(\cdot)$, the GANs is applied \cite{goodfellow2014generative} due to its powerful generating ability. In particular, we propose a two path-way network N2DGAN with a generator network $G_{\theta_{G}}$ and a discriminator network $D_{\theta_{D}}$ parameterized by $\theta_{G}$ and $\theta_{D}$, respectively.
For the discriminator network $D_{\theta_{D}}$, we will have the following maximization problem given $G_{\theta_{G}}$:
\begin{equation}
\hat{\theta}_D=\mathop{argmax}_{\theta_{D}\in\mathcal{D}}{E}_{I^{d} \sim {P}_{d}}[D_{\theta_{D}}(I^{d})]\hspace{0.1cm}-\hspace{0.1cm}{E}_{I^{n}_{t}\sim {P}_{n}}[D_{\theta_{D}}(G_{\theta_{G}}(I^{n}_{t}))]
\label{function_obj}
\end{equation}
where $P_d$ is the real daytime image distribution, and ${P}_{n}$  is the nighttime image distribution.
Here we adopt the same formulation for Eq.(\ref{function_obj}) as in WGAN \cite{arjovsky2017wasserstein},
$\mathcal{D}$ is the set of 1-Lipschitz function, and weight clipping is utilized to enforce the Lipschitz constraint.
In essence, Eq.(\ref{e1}) and Eq.(\ref{function_obj}) are the alternative optimizing of a min-max optimization problem jointly parameterized by $\theta_{G}$ and $\theta_{D}$.
\subsection{Architecture}
\label{section_model}
The overview on the network architecture of the proposed N2DGAN is shown in Fig.\ref{fig_model1}. To leverage the preserving of background scene and foreground object(s) in generating virtual daytime image, a two-pathway generator is proposed with global sub-network for maintaining background scene and $M$ local sub-networks attending to capture local foreground information.
As illustrated in Fig.\ref{fig_model1}, both the global and local sub-networks are designed in an Encoder-Decoder manner as in most cases with modules of norm Convolution-BatchNorm-Relu, and each layer is followed by three residual blocks\cite{he2016deep}.
Following the architecture of ``U-Net'' adopted in \cite{Pixel2Pixel2017}, a fusion subnet is also  designed to connect both the "Encoder" and the "Decoder" since symmetric layers can share some common information. This will be helpful for facilitating the information flow of foreground object between the input and output in the network chain. The details of the model architecture are provided in the Appendix.
\begin{figure*}[!ht]
\setlength{\belowcaptionskip}{-0.3cm}
\centering
\includegraphics[width=0.9\textwidth]{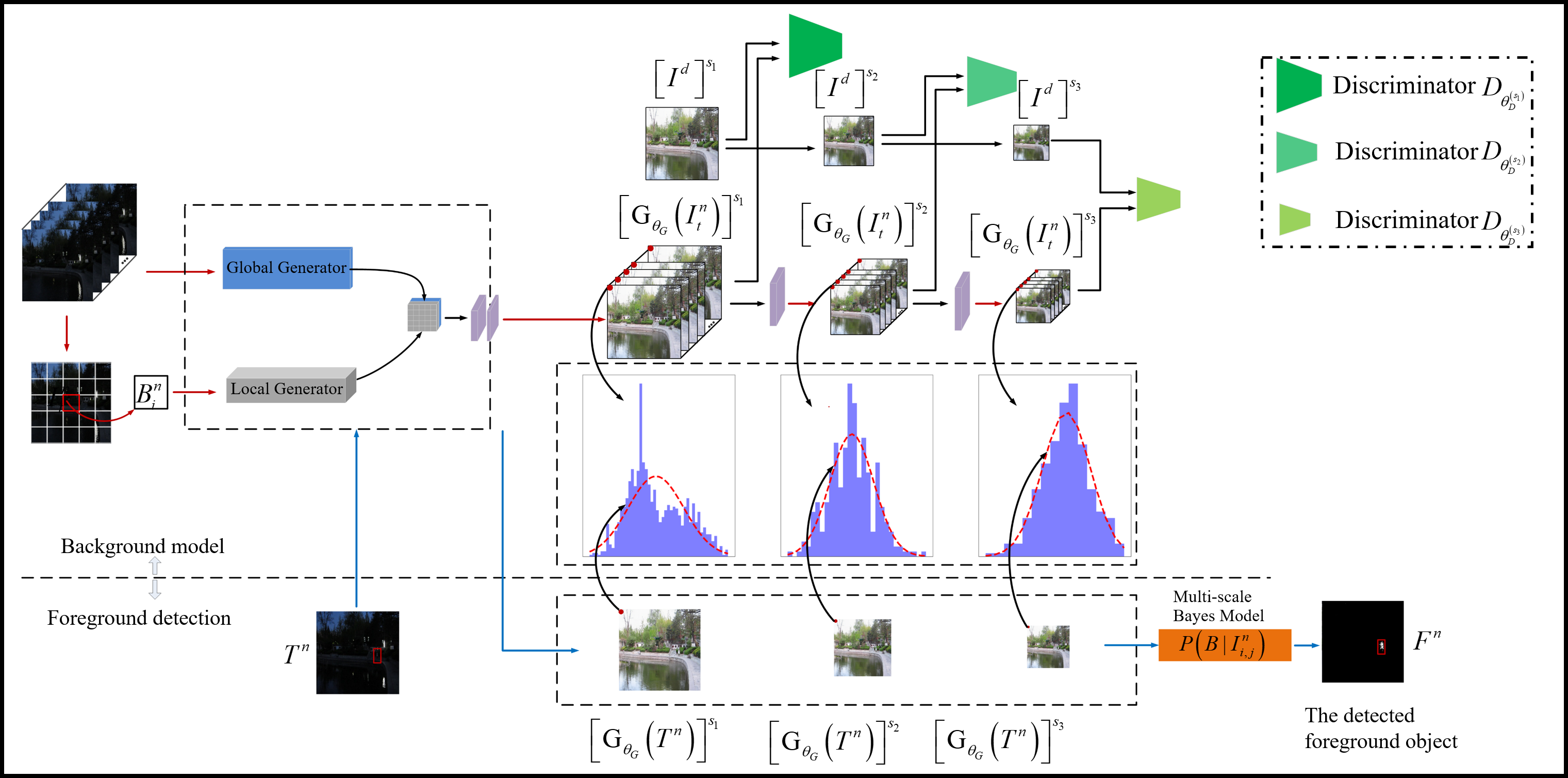}
\caption{Multi-scale Bayes inference framework for foreground detection. In the training phase, phase, for input frame $I_{t}^{n}$, multi-scale daytime images $[G_{\theta_{G}}(I_{t}^{n})]^{s_{i}}, i=1,2,...Q$, are generated, then we model the temporal distribution of each pixel of the generated image at each scale. $D_{\theta_{D}^{(s_{i})}}$ is discriminator network for scale $s_{i}$. In the testing phase, that is, foreground detection phase, for the input frame $T_{t}^{n}$, the pre-trained N2DGAN network outputs its corresponding multi-scale daytime images $[G_{\theta_{G}}(T_{t}^{n})]^{s_{i}}, i=1,2,...Q$. Then the multiple probabilities of each pixel belonging to the background at each scale are integrated elegantly based on Bayes inference for detecting the foreground object.}
\label{fig_model2}
\end{figure*}
\subsection{Loss Function}
The loss function $L(\cdot,\cdot)$ in Eq.({\ref{e1}}) plays a significant role in training GANs model. For an input nighttime image $I_{t}^{n}$, several kinds of loss functions are exploited to make the generated virtual daytime image $G_{\theta_{G}}(I_{t}^{n})$ retain most of the image information such as structure, objects, and texture as in the pre-specified reference image $I_d$ .

{\bfseries Adversarial loss.} To encourage the generated images move towards the real daytime image manifold and generate images with more details, the adversarial loss is first considered for distinguishing the generated image $G_{\theta_{G}}(I^{n})$ from the daytime image $I^d$.
\begin{equation}
L_{adv}=\frac{1}{N}\sum_{t=1}^{N}-D_{\theta _{D}}\left ( G_{\theta _{G}}\left ( I_{t}^{n} \right ) \right )
\label{function_adv}
\end{equation}
where $N$ is the number of nighttime images in the training dataset.

{\bfseries Perceptual loss.} In order to minimize the high-level perceptual and semantic differences between
$G_{\theta _{G}}\left ( I^n_t \right )$ and $I^d$ while preventing unexpected overfitting coming from $I^d$,
we follow the idea that minimizes the difference in convolutional layer of a pre-trained network \cite{gatys2016image} between two images. The motivation behind it lies in that the neural network pre-trained by image classification task has already learnt effective representation, which can be transferred into other tasks such as our enhancement processing.
Specifically, we define $\phi_i$ as the activation of the $i^{th}$ convolutional layer of the pre-trained network, and the perceptual loss is defined as:
\begin{equation}
L_{p} \!=\! \frac{1}{C} \sum_{i=1}^{C} \frac{1}{W_i} \frac{1}{H_i} \sum_{x=1}^{W_i} \sum_{y=1}^{H_i}
\left ([\phi_{i}(G_{\theta_{G}}(I^{n}_{t}))]_{x,y} - [\phi_{i}(I^{d})]_{x,y}\right )^{2}
\label{function_perceptual}
\end{equation}
where $C$ is the number of convolutional layers, $W_{i}$ and $H_{i}$ describe the dimensions of the respective feature maps within the VGG network.

{\bfseries Pixel-wise loss.}
To facilitate further background modeling task with spatial consistency of intra-frame and pixel-wise temporal consistency of inter-frame, the most widely used pixel-wise MSE loss (Eq.(\ref{function_mse})) and total variation loss (Eq.(\ref{function_tv})) are also adopted.
\begin{equation}
L_{mse}= \frac{1}{W}\frac{1}{H}\sum_{i=1}^{W}\sum_{j=1}^{H}\left ( I_{i,j}^{d}- [G_{\theta _{G}}\left ( I^{n}_{t} \right )]_{i,j}\right )^{2}
\label{function_mse}
\end{equation}
\begin{equation}
L_{tv} =\hspace{-0.2cm} \sum_{i=1,j=1}^{W, H}\hspace{-0.2cm}\sqrt{\left ((\hat{I}^d_{t})_{i+1,j}-(\hat{I}^d_{t})_{i,j}\right )^{2}+\left ((\hat{I}^d_{t})_{i,j+1} - (\hat{I}^{d}_{t})_{i,j}\right )^{2}}
\label{function_tv}
\end{equation}
where $\hat{I}^d_{t}$ denotes the generated virtual daytime image $G_{\theta_G}(I^n_{t})$.

Since each of the loss functions mentioned above is provided with an unique view on characterizing the visual quality of the generated virtual image, an intuitive way is to make a combination of them. Thus, we have the final overall loss function as:
\begin{equation}
L=\lambda _{adv}L_{adv} + \lambda_{tv}L_{tv} + \lambda_{mse}L_{mse} + \lambda_{p}L_{p}
\label{function_model1_L}
\end{equation}
where $\lambda_{adv}$, $\lambda_{tv}$, $\lambda_{mse}$, and $\lambda_{p}$ are weights of the corresponding terms, respectively.
\section{Multi-scale Bayes inference for foreground detection}
\begin{figure*}[!ht]
  \centering
  \begin{align*}
  \label{bayesian1}
  \!\!\!\!\!\!\!\! P(B \! \mid \! I_{i,j}^{n})\!=\!
  P(B \! \mid \! [G_{\theta _{G}} ( I^{n} )]_{i,j}^{s_{1}},...,[G_{\theta _{G}} ( I^{n} )]_{\left \lfloor i/2^{Q-1} \right \rfloor,\left \lfloor j/2^{Q-1} \right \rfloor}^{s_{Q}}) P([G_{\theta _{G}} ( I^{n} )]_{i,j}^{s_{1}},...,[G_{\theta _{G}} ( I^{n} )]_{\left \lfloor i/2^{Q-1} \right \rfloor,\left \lfloor j/2^{Q-1} \right \rfloor}^{s_{Q}} \! \mid \! I_{i,j}^n)
  \tag{10} 
  \end{align*}
\end{figure*}
\begin{figure*}[!ht]
  \centering
  \begin{align*}
  \label{bayesian}
  P(B\mid I_{i,j}^{n})
  &=\prod_{k=1}^{Q} P(B\mid [ G_{\theta _{G}} ( I^{n} )]_{\left \lfloor i/2^{k-1} \right \rfloor,\left \lfloor j/2^{k-1} \right\rfloor}^{s_{k}})
  P([ G_{\theta _{G}} ( I^{n} )]_{\left \lfloor i/2^{k-1} \right \rfloor,\left \lfloor j/2^{k-1} \right \rfloor}^{s_{k}} \mid I_{i,j}^n)\\
  &\propto \prod_{k=1}^{Q} P(B\mid [ G_{\theta _{G}} ( I^{n} )]_{\left \lfloor i/2^{k-1} \right \rfloor,\left \lfloor j/2^{k-1} \right\rfloor}^{s_{k}})
  \tag{11} 
  \end{align*}
\end{figure*}
N2DGAN ensures that there is a detectable difference between foreground object and background.
All these characters match the major premise of GMM, that the background is more frequently visible than the foreground and that its variance is significantly slight.
However, as we know, the neural network has the properties of both randomness and uncertainty. Thus, there exists inevitably pixel-wise difference between $I_{i,j}^{d}$ and $[G_{\theta _{G}}( I^{n})]_{i,j}$ in generating $G_{\theta _{G}}( I^{n})$, and it essentially can be regarded as some kind of random noise arising from both spatial and temporal domains. In other words, given the total error value, this difference at pixel $(i,j)$ may also occur at any other pixels with equal probability. This case will be doomed to bring some unexpected negative influence on pixel-level background modeling.

To mitigate this issue, inspired by some works on multi-scale multiplication for edge detection \cite{rosenfeld1970edge,rosenfeld1971edge} that tend to yield significant localized detection, we extend N2DGAN to a multi-scale generative model as shown in Fig.\ref{fig_model2} to facilitate the background modeling to be noise-free.

\subsection{Multi-scale Generation}
As illustrated in Fig.\ref{fig_model2}, we reformulate the generation problem for background modeling as follows:
to train a generator network $G_{\theta_{G}}$ parametrized by $\theta_{G}$, which learns a mapping  function from the source domain $\Omega_n$ of nighttime to the target domain $\Omega_d$ of daytime. For every input nighttime frame $I_{t}^{n}\in \Omega_n$,
to generate a multi-scale set of images $[G_{\theta _{G}}(I_{t}^{n})]^{s_{i}}$, $i=1,......,Q$, in daytime domain $\Omega_d$ will be equivalent to:
\begin{equation}
\hat{\theta_{G}} = \frac{1}{N}\mathop{arg min}_{\theta_{G}}\sum_{t=1}^{N}\sum_{i=1}^{Q} L([G_{\theta_{G}}(I_{t}^{n})]^{s_{i}},[I^{d}]^{s_i})
\label{function_model2_1}
\end{equation}
where $N$ is the number of training pairs as before, and $L(\cdot,\cdot)$ denotes the loss function as mentioned above. In addition, we introduce $Q$ adversarial discriminators to distinguish the reference daytime image $[I^{d}]^{s_{i}} $ from the generated virtual daytime image $[G_{\theta _{G}}(I_{t}^{n})]^{s_{i}}$ under scale $s_{i}$.
Particularly, similar to Eq.({\ref{function_obj}}), in each generation task of different scales, the discriminator network will be reformulated as :
\begin{equation}
\begin{split}
\hat{\theta}_{D}^{(s_{i})}=\mathop{argmax}_{\theta_{D}^{(s_{i})} \in \mathcal{D}} &{E}_{[I^{d}]^{s_{i}} \sim {P}_{d}}[D_{\theta_{D}^{(s_{i})}}([I^{d}]^{s_{i}})] \\
&- {E}_{I^{n} \sim {P}_{n}}[D_{\theta_{D}^{(s_{i})}}([G_{\theta_{G}}(I^{n})]^{s_{i}})]
\end{split}
\label{function_model2_2}
\end{equation}

It should be noted that both the generative network architecture and the discriminator architecture in Fig.\ref{fig_model2} are same as those in Fig.\ref{fig_model1}. But different from N2DGAN, more convolutional layers are employed to generate multi-scale daytime images.
\subsection{Multi-scale Bayes Inference Based on Scale Multiplication}
N2DGAN enforces the sequence of generated virtual daytime images $G_{\theta_{G}}(I^{n}_t)$, $t=1,...,N$, to be as approximated closely as possible to the pre-specified unique reference frame $I^d$.
The inescapable fact, however, is that there is certain difference between them, one is noise accompanied by the neural network, the other part is the foreground region. To suppress effectively noise coming from virtual daytime generation while strengthening the discriminant of foreground object, a multi-scale Bayes model is proposed to characterize pertinently the temporal variation of background.

For each pixel $I_{i,j}^n$, we use $P(B\mid I_{i,j}^{n})$ to serve as the background model, denoting the probability of pixel $I_{i,j}^n$ to be background. Given the multi-scale representations [$G_{\theta _{G}} ( I^{n} )]_{i,j}^{s_{k}},k=1,...,Q,$ for pixel $ I_{i,j}^n$ ,
the background model $P(B\mid I_{i,j}^{n})$ can be given with Bayes criterion by Eq.(\ref{bayesian1}). Here, $\left \lfloor \cdot \right \rfloor$ represents rounding down to the nearest whole number. On the assumption that the generation of virtual daytime images with different scales is independent of each other, thus the background model given by Eq.(\ref{bayesian1}) will further reduce to the following Eq.(\ref{bayesian}). As we can see from Eq.(\ref{bayesian}), the background model $P(B\mid I_{i,j}^{n})$ is equivalent to the multi-scale multiplication of multiple background models at different scales. In addition, for each background model $P(B\mid[G_{\theta _{G}} ( I^{n} )]_{\left \lfloor i/2^{k-1} \right \rfloor,\left \lfloor j/2^{k-1} \right\rfloor}^{s_{k}})$, $k=1,...,Q$, a single gaussian model instead of GMM can be simply applied as shown in Fig.\ref{fig_model2}.
\section{Experimental Results}
We evaluate the proposed background modeling approach visually and quantitatively, by comparing with state-of-the-arts and providing extensive ablation studies
\subsection{Datasets and Experiment Settings}
{\bfseries Datasets and evaluation metrics.}
Our work in this paper mainly focuses on background modeling under nighttime scene with low illumination. However, to the best of our knowledge, there are no public open datasets to evaluate such a task. For this reason, we collect several benchmark datasets by a Canon IXY 210F video camera including indoor and outdoor scenes with manually labeled ground truth. The details about our four datasets, including \textbf{Lab}, \textbf{Tree}, \textbf{Lake1}, and \textbf{Lake2}, are shown in Tab.\ref{table:datasets}
~\footnote{The datasets and code of our method will be released at https://github.com/anqier0468/N2DGAN.}
. For each dataset, the corresponding pre-specified daytime images that serve as ground truth background frames are also provided. It is worth of noting that both the `Lake' and `Tree' datasets were captured outdoor on windy nights, and the 'Lab' dataset is taken indoors where we control the intensity of lighting by pulling curtains and switching incandescent lights on purpose. Actually, these datasets are much challenging for background modeling task since they feature the undulation of lake, reflection of lights in the water, leaves shaking, and illumination variation. In order to make a quantitative evaluation, the foreground object(s) in the datasets are also manually labeled. Following the previous works on foreground detection, {\textbf{IoU}} is employed as our evaluation metric~\cite{liu2012foreground}.

{\bfseries Implementation details.} The training of the proposed N2DGAN model is implemented on $2$ NVIDIA TITAN Xp GPUs.
The first $300$ frames of each nighttime video in Tab.\ref{table:datasets} paired with the corresponding daytime reference image are used to train the model, and the remaining are for testing.
All of the images are downscaled to resolution of $256\times256$. Specially, we split each image into multiple image blocks with size $32\times32$, and then each block is used as the input of each local generator subnet. Considering the computational efficiency, only two scales are adopted, i.e., $Q=2$, to eliminate the influence of noise and spatial shift of background pixels. Based on RMSProp, the mini-batch gradient descent method is used with a batch size of $4$ and a learning rate of $10^{-4}$. Since WGAN~\cite{arjovsky2017wasserstein} is used as the backbone of our generation model, the weight need to be clampped to a fixed box $(-0.01, 0.01)$ after each gradient update to avoid gradient vanishing and mode collapse problems during the learning process. In all our experiments, we uniformly set the total training epoch number to 30, and empirically set $\lambda _{adv} = 10^{-1}$, $\lambda _{p} = 10^{-5} $, $ \lambda _{mse} = 10^{-1}$ , and $\lambda _{tv} = 10^{-3} $ in Eq.(\ref{function_model1_L}) to maintain the same order of magnitude. Besides, the first $3$ convolutional layers of VGG network is used to calculate perceptual loss.
\begin{table}[!h]

	\centering
    \caption{The details of our four datasets. }
    \begin{tabular}{c|c|c|c|c} 
      \hline
      \hline
      Datasets & Lab & Tree & Lake1 & Lake2\\
      \hline
      \#Frames &693&722&579&1482\\
      \hline
      {\tabincell{c}{Nighttime\\Scene}}&
      \mgape{\includegraphics[width=0.08\textwidth]{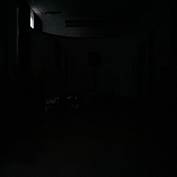}} & \mgape{\includegraphics[width=0.08\textwidth]{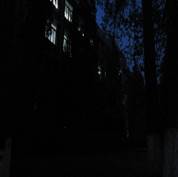}} & \multicolumn{2}{c}{\mgape{\includegraphics[width=0.08\textwidth]{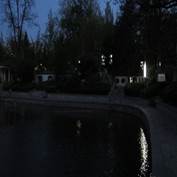}}}\\
      \hline
       {\tabincell{c}{Reference\\Image}}&
      \mgape{\includegraphics[width=0.08\textwidth]{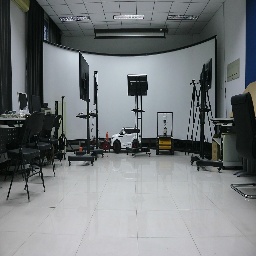}} & \mgape{\includegraphics[width=0.08\textwidth]{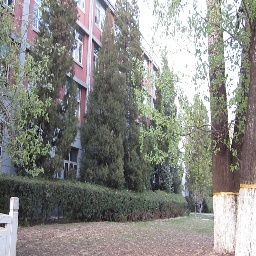}} & \multicolumn{2}{c}{\mgape{\includegraphics[width=0.08\textwidth]{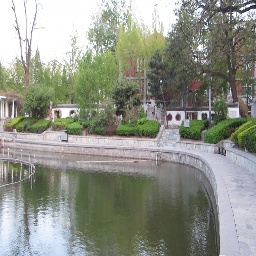}}}\\
      \hline
      \hline
      \end{tabular}
\label{table:datasets}%
\end{table}
\begin{figure*}[!ht]
  \centering
  \includegraphics[width=1.00\textwidth]{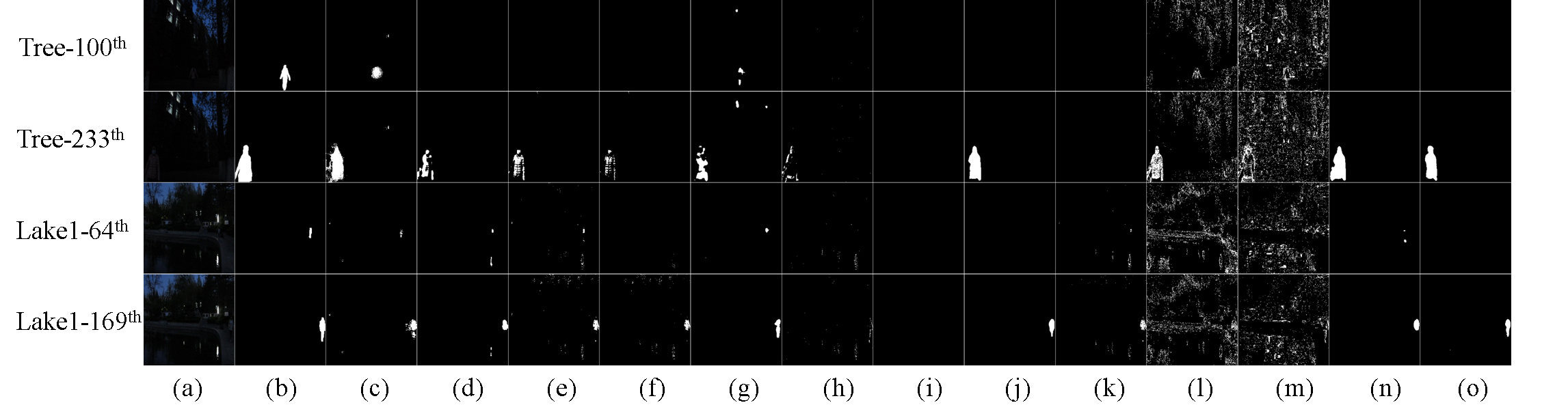}
  \caption{Qualitative comparison of different foreground detection methods. (a) nighttime image, (b) groundtruth, (c) our method N2G-GAN, (d) GMG \cite{godbehere2012visual}, (e) ASOM \cite{maddalena2008self}, (f) FASOM \cite{maddalena2010fuzzy}, (g) LOBSTER \cite{st2014improving}, (h) GMM \cite{Zivkovic2004Improved}, (i) MCueBGS \cite{Noh2012A}, (j) SUBSENSE \cite{st2014flexible}, (k) MRF-UV \cite{zhao2012fuzzy}, (l) HE \cite{Ketcham1976Real}+GMM \cite{Zivkovic2004Improved}, (m) MSR \cite{Jobson2002A}+GMM \cite{Zivkovic2004Improved}, (n) HE \cite{Ketcham1976Real}+SUBSENSE \cite{st2014flexible}, (o) MSR \cite{Jobson2002A}+SUBSENSE \cite{st2014flexible}.} 
  \label{fig_compareWithTradition_lake}
\end{figure*}
\subsection{Performance Evaluation}
{\bfseries Comparisons with State-of-the-arts.} We compare the proposed method with state-of-the-arts on foreground detection, including {\bfseries{{eight}}} typical background modeling methods and {\bfseries{{four}}} enhancement-based methods.
Implementations of all these methods are based on the BGSLibrary \cite{bgslibrary} with default parameters.
The quantitative comparison results on $3$ sequences are shown in Tab.\ref{tableDetectionAccuracy}. Obviously, our method can always achieve the best performance. Specifically, for sequences 'Tree' and 'Lake2', our method even outperforms the state-of-the-art method SUBSENSE \cite{st2014flexible} by 75\% and 44\%, respectively.
Fig.\ref{fig_compareWithTradition_lake} presents the qualitative comparison results, which illustratively show that the proposed N2DGAN performs better.
This mainly lies in two facts: 1) compared with the directly background modeling methods, generating daytime images makes the flatten pixel distribution sharper and easier to detect foreground; 2) compared with the enhancement-based methods, the unique daytime reference frame in generative process ensures the inter-frame consistency of the generated daytime images.
\begin{figure}[htbp]
\setlength{\belowcaptionskip}{-0.cm}
    \centering
        \includegraphics[width=0.459\textwidth]{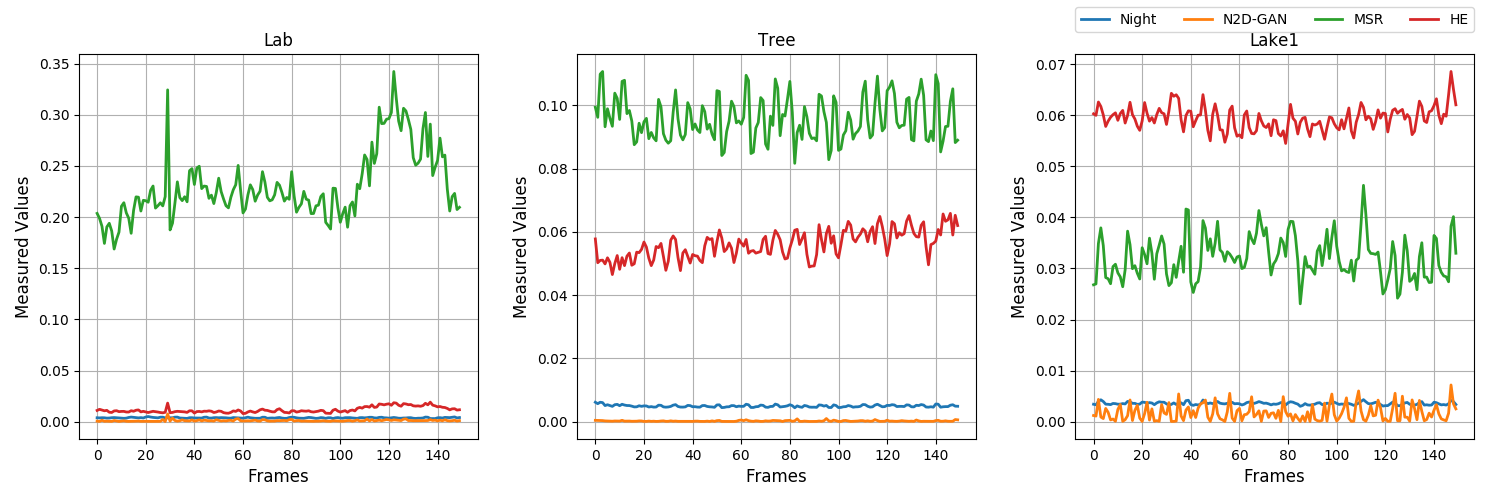}
    \caption{Sequence stability comparison with HE, MSR and N2DGAN on Lab (left), Tree (middle), and Lake1 (right) datasets
  }
  \label{fig_stabilityKL}
\end{figure}


\begin{table*}[!t]
\centering
    \caption{Quantitative comparison of foreground detection accuracy by different methods.}
    \begin{tabular}{c|c|c|c|c} 
      \hline
      \hline
      \multicolumn{2}{c|}{\multirow{2}*{Methods}}&\multicolumn{3}{c}{Accuracy($\%$)}\\
      \cline{3-5}
      \multicolumn{2}{c|}{~} &{Tree}&{Lake1}&{Lake2}\\
      \hline
      \multirow{9}*{Typical methods}&GMG\cite{godbehere2012visual}&$13.07\%\pm2.97\%$ & $21.45\%\pm2.03\%$ & $10.70\%\pm1.30\%$\\
      &GMM\cite{Zivkovic2004Improved}&$5.85\%\pm0.47\%$&$6.90\%\pm0.16\%$&$7.19\%\pm0.25\%$\\
      &ASOM\cite{maddalena2008self}&$10.01\%\pm1.93\%$&$16.41\%\pm1.49\%$&$18.40\%\pm0.80\%$\\
      &FASOM\cite{maddalena2010fuzzy}&$4.42\%\pm0.53\%$&$13.55\%\pm1.61\%$&$10.16\%\pm0.27\%$\\
      &IMBGS\cite{bloisi2012independent}&$12.13\%\pm2.69\%$ & $10.98\%\pm3.74\%$ & $10.46\%\pm2.16\%$\\
      &MRF-UV\cite{zhao2012fuzzy}&$16.43\%\pm4.71\%$&$17.36\%\pm2.32\%$&$8.82\%\pm0.60\%$\\
      &LOBSTER\cite{st2014improving}&$26.89\%\pm2.60\%$&$37.87\%\pm3.89\%$&$22.23\%\pm3.64\%$\\
      &SUBSENSE\cite{st2014flexible}&$28.43\%\pm7.93\%$&$25.89\%\pm7.60\%$&$26.73\%\pm3.42\%$\\
      \hline
      \multirow{4}*{\tabincell{c}{Enhancement-based\\methods}}
      &HE\cite{Ketcham1976Real}+GMM\cite{Zivkovic2004Improved}&$13.08\%\pm1.45\%$&$2.02\%\pm0.02\%$&$2.93\%\pm0.12\%$\\
      &HE\cite{Ketcham1976Real}+SUBSENSE\cite{st2014flexible}&$33.95\%\pm8.09\%$&$22.89\%\pm2.66\%$&$22.33\%\pm2.88\%$\\
      &MSR\cite{Jobson2002A}+GMM\cite{Zivkovic2004Improved}&$6.53\%\pm0.33\%$&$2.21\%\pm0.03\%$&$2.73\%\pm0.07\%$\\
      &MSR\cite{Jobson2002A}+SUBSENSE\cite{st2014flexible}&$29.05\%\pm7.25\%$&$19.99\%\pm4.29\%$&$17.24\%\pm2.02\%$\\
      &LIME\cite{DBLP:Guo2016}+GMM\cite{Zivkovic2004Improved}&$8.91\%\pm1.45\%$&$6.18\%\pm0.17\%$&$5.69\%\pm0.35\%$\\
      &LIME\cite{DBLP:Guo2016}+SUBSENSE\cite{st2014flexible}&$17.73\%\pm6.59\%$&$38.46\%\pm7.37\%$&$30.74\%\pm4.60\%$\\
      &LLNet\cite{DBLP:Lore2017}+GMM\cite{Zivkovic2004Improved}&$6.11\%\pm0.72\%$&$12.74\%\pm0.37\%$&$8.87\%\pm0.39\%$\\
      &LLNet\cite{DBLP:Lore2017}+SUBSENSE\cite{st2014flexible}&$19.03\%\pm7.60\%$&$37.17\%\pm3.06\%$&$25.51\%\pm2.90\%$\\
      &Struct\cite{DBLP:Li2018}+GMM\cite{Zivkovic2004Improved}&$2.85\%\pm0.20\%$&$6.44\%\pm0.14\%$&$4.81\%\pm0.14\%$\\
      &Struct\cite{DBLP:Li2018}+SUBSENSE\cite{st2014flexible}&$16.00\%\pm6.33\%$&$39.88\%\pm2.60\%$&$27.14\%\pm3.37\%$\\
      \hline
      \multirow{3}*{Ours}&N2DGAN(global)&$20.61\%\pm2.72\%$&$35.74\%\pm1.82\%$&$22.91\%\pm1.52\%$\\
      &N2DGAN(local)&$36.46\%\pm1.82\%$&$37.61\%\pm0.93\%$&$27.68\%\pm1.03\%$\\
      \cline{2-5}
      &N2DGAN&\bm{${48.45\%}\pm{1.93\%}$} &\bm{ ${ 40.32\%}\pm{ 1.69\%}$} & \bm{ ${ 39.29\%}\pm{ 1.91\%}$}\\
      \hline
      \hline
      \end{tabular}
\label{tableDetectionAccuracy}%
\end{table*}

{\bfseries Stability comparison.} Consistency stability of successive frames is of great importance for further background modeling. For the $t$-th frame $I_{t}\epsilon R^{W\times H}$, we use the following metric to measure the stability between $I_{t}$ and its adjacent frame $I_{t+1}$.
\begin{equation}
\setcounter{equation}{12}
s_t = Sim(I_t,I_{t+1})
\label{function_stability}
\end{equation}
where $Sim(\cdot,\cdot)$ denotes the distance between $I_t$ and $I_{t+1}$. Here, we adopt Kullback-Leibler Divergence, which represents stronger stability if its value is close to $0$.
Fig.\ref{fig_stabilityKL} shows the stability comparison of several representative methods on a randomly selected sequence consisting of 200 consecutive frames from the test set of three datasets.
As we can see, the result of our method (yellow line) is quite close to real nighttime images (blue line), while both HE (red line) and MSR (green line) show distinct difference from real sequence. Particularly, the large fluctuation by MSR also indicates that the pixel values between two adjacent frames differ greatly, which is unfavorable for background modeling. To sum up, with the joint constrain of spatial and temporal consistency, our generation based method performs better than enhancement-based method.
\begin{figure}[htbp]
\centering
\includegraphics[width=0.4\textwidth]{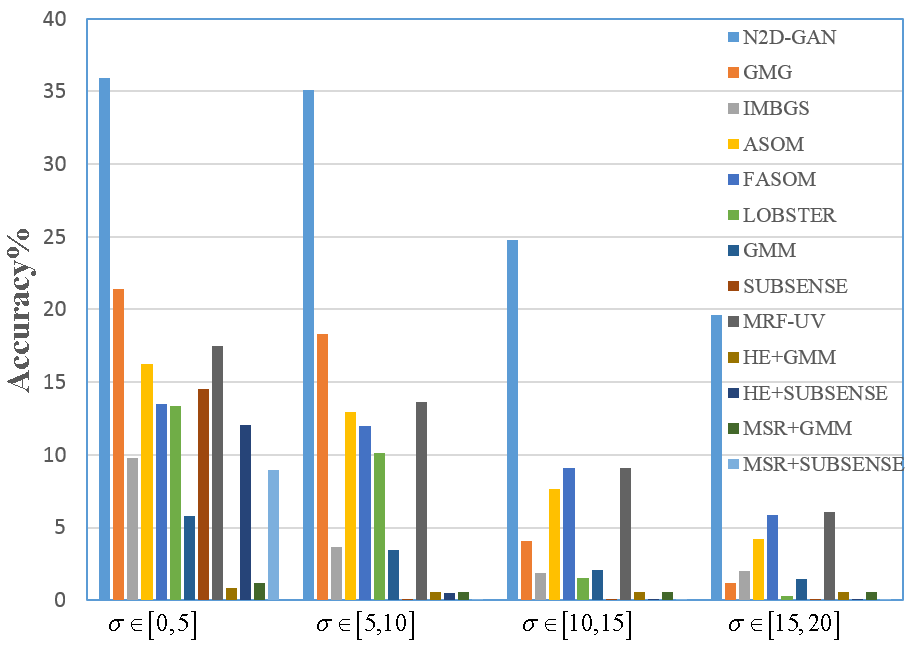}
\caption{Foreground detection accuracy on Lake1 dataset with different levels of noises.}
\label{fig_noiseLine}
\end{figure}
\subsection{Robustness to Noise and Illumination Variation}
\begin{figure*}[!ht]
  \centering
  \includegraphics[width=0.99\textwidth]{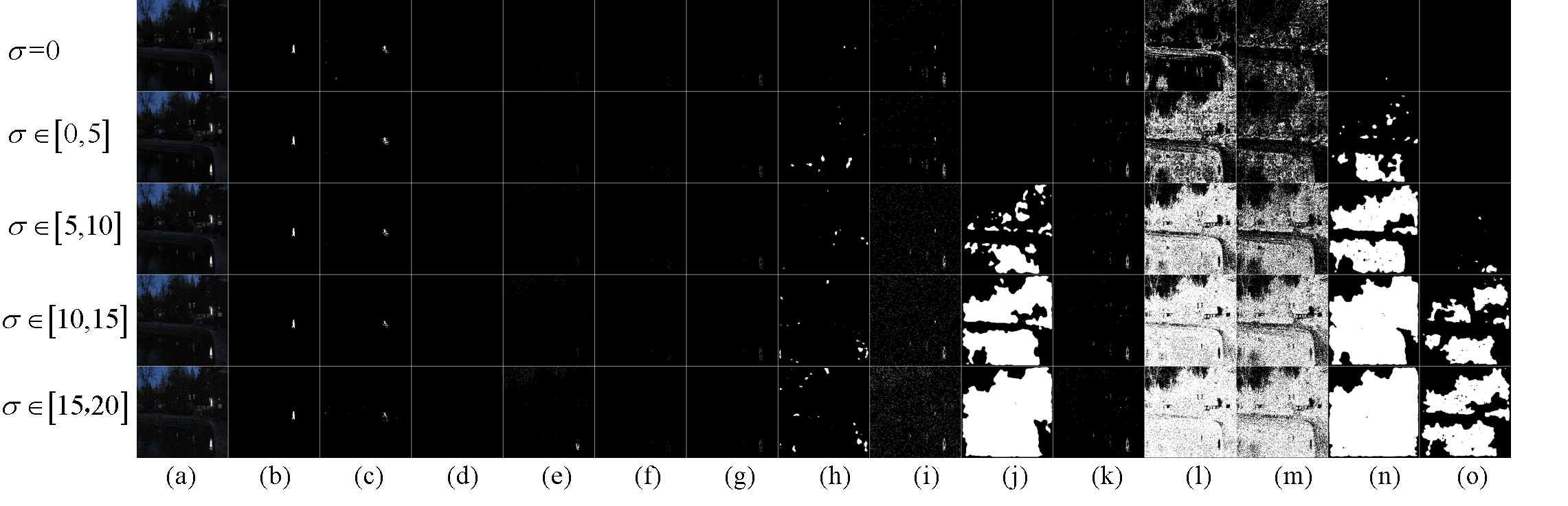}  
  \caption{Performance evaluation of robustness to noise on Lake1-13th.
    (a) The input nighttime image.
    (b) Groundtruth.
    (c) N2DGAN.
    (d) GMG\cite{godbehere2012visual},
    (e) IMBGS\cite{bloisi2012independent}.
    (f)ASOM \cite{maddalena2008self},
    (g)FASOM \cite{maddalena2010fuzzy},
    (h)LOBSTER \cite{st2014improving},
    (i)GMM \cite{Zivkovic2004Improved},
    (j)SUBSENSE \cite{st2014flexible},
    (k)MRF-UV \cite{zhao2012fuzzy},
    (l) HE \cite{Ketcham1976Real}+GMM \cite{Zivkovic2004Improved}, (m) MSR \cite{Jobson2002A}+GMM \cite{Zivkovic2004Improved}, (n) HE \cite{Ketcham1976Real}+SUBSENSE \cite{st2014flexible}, (o) MSR \cite{Jobson2002A}+SUBSENSE \cite{st2014flexible}.}
  \label{fig_compareNoise1}
\end{figure*}

\begin{figure*}[!ht]
  \centering
  \includegraphics[width=0.99\textwidth]{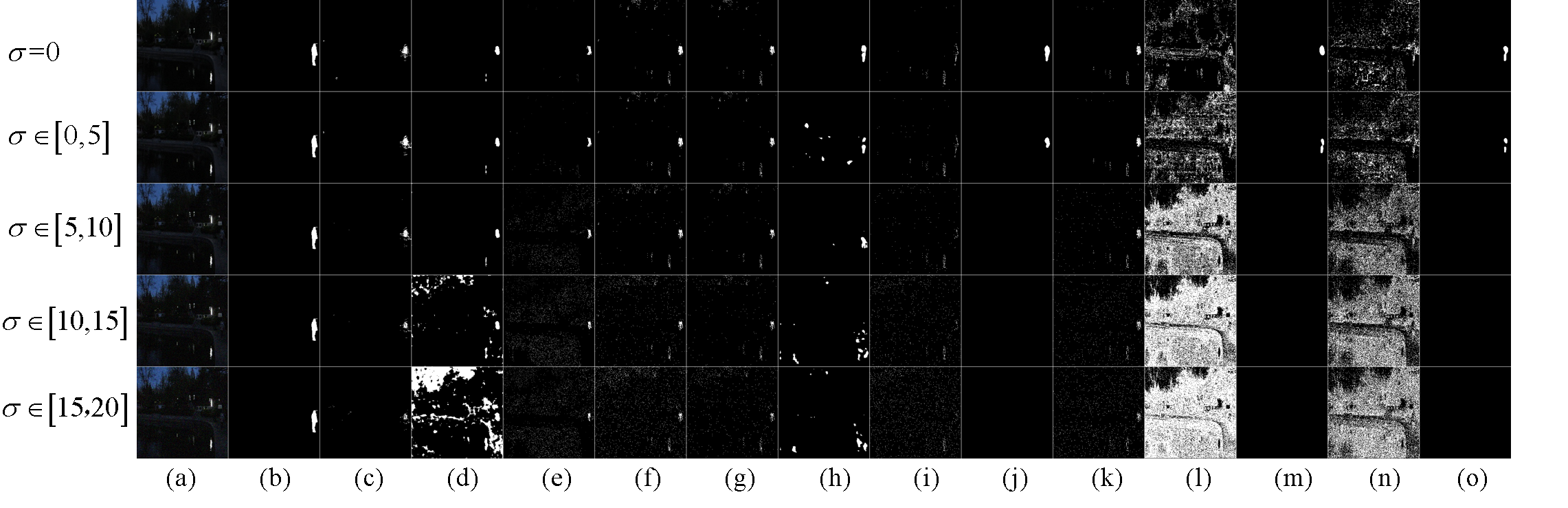}  
  \caption{Performance evaluation of robustness to noise on Lake1-158th.
    (a) The input nighttime image.
    (b) Groundtruth.
    (c) N2DGAN.
    (d) GMG\cite{godbehere2012visual},
    (e) IMBGS\cite{bloisi2012independent}.
    (f)ASOM \cite{maddalena2008self},
    (g)FASOM \cite{maddalena2010fuzzy},
    (h)LOBSTER \cite{st2014improving},
    (i)GMM \cite{Zivkovic2004Improved},
    (j)SUBSENSE \cite{st2014flexible},
    (k)MRF-UV \cite{zhao2012fuzzy},
    (l) HE \cite{Ketcham1976Real}+GMM \cite{Zivkovic2004Improved}, (m) MSR \cite{Jobson2002A}+GMM \cite{Zivkovic2004Improved}, (n) HE \cite{Ketcham1976Real}+SUBSENSE \cite{st2014flexible}, (o) MSR \cite{Jobson2002A}+SUBSENSE \cite{st2014flexible}.}
  \label{fig_compareNoise2}
\end{figure*}

Some extreme weather conditions such as rain, snow, and fog usually bring great challenges to background modeling. To demonstrate N2DGAN's scalability under such environment,
additive Gaussian noise is added to nighttime video sequence to simulate extreme weather.
We randomize the noise standard deviation $\sigma \epsilon \left \{ \left [ 0,5 \right ],\left [ 5,10 \right ],\left [ 10,15 \right ],\left [ 15,20 \right ] \right \}$ separately for each testing example.
As illustrated in Fig.\ref{fig_noiseLine},
the behaviors are quantitatively different in all three datasets. This demonstrates that our method is the only technique that manages to perform well with different levels of noises. On two randomly selected frames Lake1-13th and Lake1-158th, Fig. \ref{fig_compareNoise1} and Fig. \ref{fig_compareNoise2} present visually the comparison results with conventional background modeling methods and enhancement based methods, which clearly shows that the proposed N2DGAN achieves the best performance.

\begin{figure*}[!ht]
    \centering
    \includegraphics[width=0.99\textwidth]{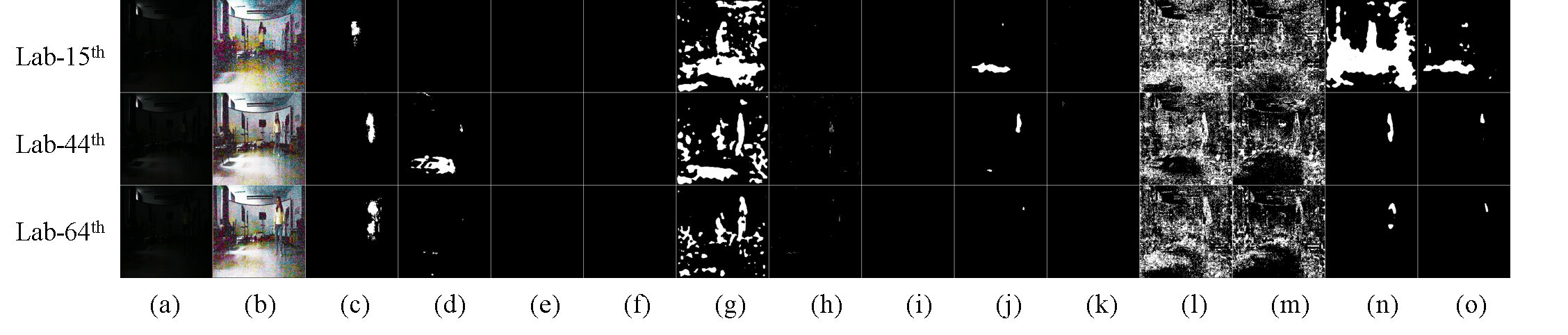}

     \caption{Performance evaluation of robustness to illumination variation. (a) nighttime image, (b) HE, (c) N2DGAN, (d) GMG \cite{godbehere2012visual}, (e) ASOM \cite{maddalena2008self}, (f) FASOM \cite{maddalena2010fuzzy}, (g) LOBSTER \cite{st2014improving}, (h) GMM \cite{Zivkovic2004Improved}, (i) MCueBGS \cite{Noh2012A}, (j) SUBSENSE \cite{st2014flexible}, (k) MRF-UV \cite{zhao2012fuzzy}, (l) HE \cite{Ketcham1976Real}+GMM \cite{Zivkovic2004Improved}, (m) MSR \cite{Jobson2002A}+GMM \cite{Zivkovic2004Improved}, (n) HE \cite{Ketcham1976Real}+SUBSENSE \cite{st2014flexible}, (o) MSR \cite{Jobson2002A}+SUBSENSE \cite{st2014flexible}.}
    \label{fig_illumination}
\end{figure*}

For the experiments on evaluating the robustness to illumination variation, Fig. \ref{fig_illumination} illustrates the comparison results on the indoor nighttime dataset Lab with illumination variation. As we can see from Fig.\ref{fig_illumination}, the N2DGAN model is much more insensitive to the instantaneous changes in light compared with other state-of-the-art background modeling methods. Here, the enhancement result based on HE (Fig. \ref{fig_illumination}(b)) is only utilized to clarify the foreground object since it is not easy to find the ground truth in dark.


Two factors must be credited for our high resilience to noise and illumination change. The first originates from our model design, which allows noisy pixel to be outfitting to the reference background image.
The second lies in our background model on successive multi-scale images, which is more robust to noise by hierarchical Bayes modeling.


\subsection{Ablation Study}
\begin{figure}[htbp]
  \centering
  \includegraphics[width=0.45\textwidth]{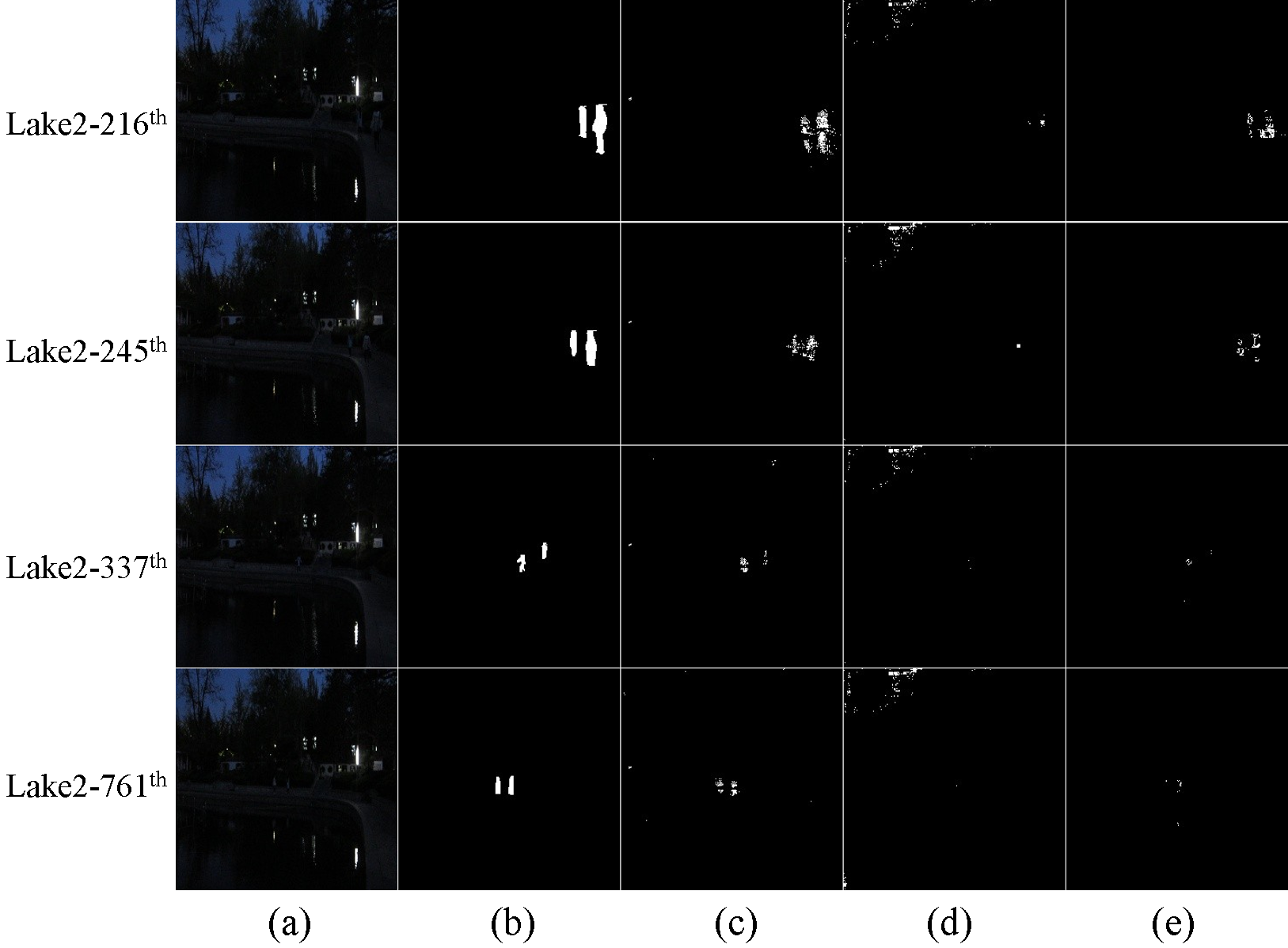}
  \caption{Comparison of foreground detection results using global subnet and local subnet independently.
  (a) Input nighttime image.
  (b) Ground truth.
  (c) N2DGAN.
  (d) N2DGAN(global).
  (e) N2DGAN(local).
  }
  \label{fig_globalAndLocalConsis}
\end{figure}

{\bfseries Global and Local Consistency Evaluation.}
To verify the effectiveness of combining both local and global consistency together in our model, we first perform foreground detection when using global subnetwork alone, called {\bfseries{\emph{N2DGAN(global)}}}. As observed from Fig.\ref{fig_globalAndLocalConsis}, small targets in nighttime images are lost in this case. Meanwhile, when using local subnet alone, called {\bfseries{\emph{N2DGAN(local)}}}, the detected foreground objects are incomplete on the edge of patches, since there exists blocking-artifact problem caused by patch enhancement. Additionally, quantitative comparison results shown in Tab.\ref{tableDetectionAccuracy} (bottom) demonstrate that our baseline improves detection accuracy by more than 10\% and 5\% compared with N2DGAN(global) and N2DGAN(local), respectively.

\begin{figure}[htbp]
  \centering
  \setlength{\belowcaptionskip}{0cm}
  \includegraphics[width=0.45\textwidth]{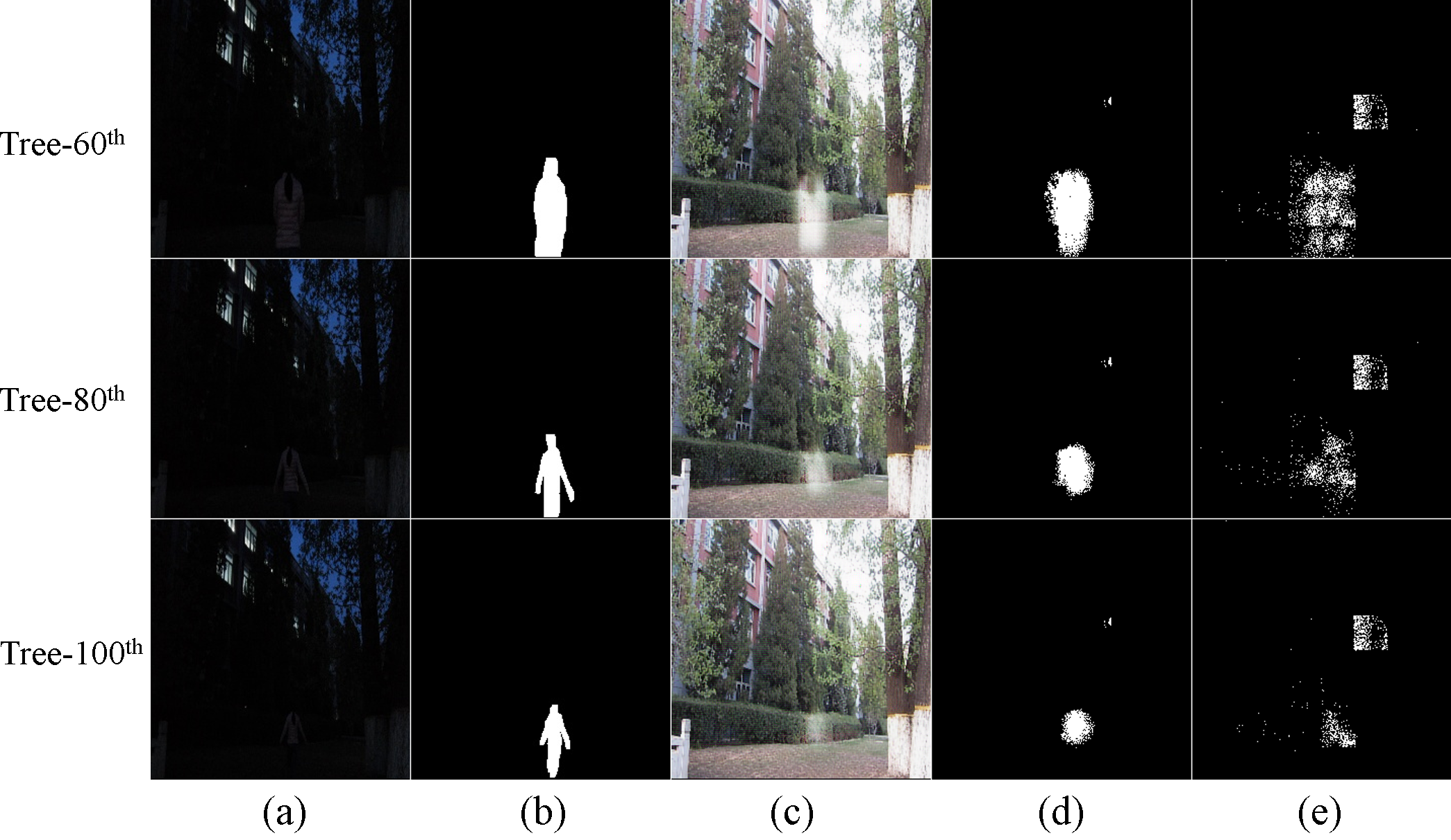}
  \caption{Comparison of foreground detection results using multi scales and single sale.
(a) nighttime image.
(b) Groundtruth.
(c) Generated image $[G_{\theta _{G}}(I_{n})]^{s_{1}}$.
(d) N2DGAN.
(e) The detection result using single scale $s_{1}$.}
\label{fig_multiScaleVSPatchBased}
\end{figure}

{\bfseries Evaluation on Multi-scale Bayes modeling .}
To further demonstrate the effectiveness of our background modeling on successive multi-scale images of daytime domain, we attempt to perform on a single scale generated images $[G_{\theta _{G}}(I_{n})]^{s_{1}}$. As illustrated in Fig.\ref{fig_multiScaleVSPatchBased}, due to the fact that multi-scale bayes model can suppress the noise caused by network, then our baseline N2DGAN makes the foreground region more remarkable.
\subsection{Time Complexity Analysis}
For a background model, the computational complexity is one of the key issues worthy of attention. For our N2DGAN model, its computational complexity mainly consists of two parts, i.e., virtual daytime image generation and foreground detection.
In the foreground detection stage, since our model holds only a single gaussian model which can be off-line available and without need for online model updating, thus the time complexity in this stage is much lower than the traditional GMM and can be negligible compared with the one in generation stage.

By comparison, to generate a virtual daytime image will occupy most of the time with a frame rate of $8$ fps without any code optimisation.
As shown in Table \ref{table_time}, the frame rate of \textbf{N2DGAN(local)} to generate $8\times8$ blocks of local sub-images\footnote{For an input nighttime image, it is divided into $8\times8$ blocks and then the corresponding local generation sub-network is trained on each block to generate a local sub-image. For details please refer to Section III.} is around $10$ fps, which is much more slowly than \textbf{N2DGAN(global)} with a frame rate of 56 fps. However, considering that we can generate each block of local virtual sub-image in parallel, the frame rate of {\textbf{N2DGAN(local)}} will dramatically increased, approximating around $640$ fps in an ideal situation. It means that the global generation process with $56$ fps will dominate the overall computational complexity of the virtual daytime image generation. By this way, the need for online real-time foreground object detection can be met.

\begin{table}[htbp]
\center
\caption{ Time complexity analysis of generation process.}
\begin{tabular}{c|c|c}
  \hline
  \hline
  \multicolumn{3}{c}{Time Complexity (fps)}\\
  \hline
  N2DGAN& N2DGAN(global)&N2DGAN(local) \\
  \hline
  8&56&10 \\
  \hline
  \hline
\end{tabular}
\label{table_time}
\end{table}
\section{Conclusion}
For the challenge of background modeling under daytime scene, an innovative N2DGAN model is proposed, which paves a new way completely different from the existing methods. To the best of our knowledge, this is the first time to introduce GANs based deep learning for this practical problem.
As an unsupervised model, N2DGAN is provided with good scalability and practical significance.

As for the time complexity of N2DGAN, it takes about $0.125$ seconds ($8$ fps) for each frame.
Considering each local generation model can be implemented in parallel, the proposed N2DGAN could be highly parameterizable.
Besides, some model compression works\cite{jia2017improving, yu2017compressing} can also be feasible solutions for network acceleration. It is also worth noting that we have assumed that the images in the training set is free of foreground objects. To overcome this limitation, we will try to extend this method to deal with the situation that the images in the training set may contain some foreground objects.


\section*{appendix}
The detailed structures of the global sub-network and local sub-network are provided in Table \ref{table_local} and Table \ref{table_global}, respectively. Each convolution layer is followed by $3$ residual block \cite{he2016deep}.
Then we simply concatenate the output from each local generator and the global generator to produce a fused feature tensor and then feed it to the successive convolution layers to generate the final output.

\begin{table}[!ht]
	\centering
	\caption{Architecture of the local generator sub-network.}
    \label{table_local}       
	\begin{tabular}{c|c|c|c}
		\hline
     	Layer & Inputs&Kernel/Stride &Outputs  \\
		\hline
		cov0&$B^{n}_{1}$&$3 \times 3 \times32/1$ & $32 \times 32 \times 32$\\
		\hline
		cov1&cov0&$3 \times 3 \times 64 / 2$ & $16 \times 16 \times 64$\\
		\hline
        cov2&cov1& $3 \times 3 \times 128/2$ & $8\times8\times128$\\ \hline
        decov0&cov2& $3 \times 3 \times 64/2$ & $16\times16\times64$\\ \hline
        decov1&cov0& $3 \times 3 \times 32/2$ & $32\times32\times32$\\ \hline
        decov2&cov2& $3 \times 3 \times 128/2$ & $16\times16\times128$\\ \hline
        decov3&\tabincell{c}{decov0,\\cov1,\\decov2}& $3 \times 3 \times 256/2$ & $32\times32\times256$\\ \hline
        decov4&\tabincell{c}{decov3,\\cov0,\\decov1}& $3 \times 3 \times 64/2$ & $F^{d}_{1}(32\times32\times64)$\\ \hline
	\end{tabular}%
\end{table}

\begin{table}[!ht]
	\centering
	\caption{Architecture of the global generator sub-network.}
    \label{table_global}       
	\begin{tabular}{c|c|c|c}
		\hline
     	Layer & Inputs&Kernel/Stride &Outputs  \\
		\hline
		cov0&$I^{n}_{t}$&$3 \times 3 \times32/1$ & $256 \times 256 \times 32$\\
		\hline
		cov1&cov0&$3 \times 3 \times 64 / 2$ & $128 \times 128 \times 64$\\
		\hline
		cov2&cov1&$3 \times 3 \times 128 / 2$ & $64 \times 64 \times 128$\\
        \hline
		cov3&cov2&$3 \times 3 \times 256 / 2$ & $32 \times 32 \times 256$\\
		\hline
		cov4&cov3&$3 \times 3 \times 512 / 2$ & $16 \times 16 \times 512$\\
		\hline
        cov5&cov4& $3 \times 3 \times 1024/2$ & $8\times8\times1024$\\
        \hline
		fc1&cov5 &- & $512$\\
		\hline
        fc2&fc1&-&$8\times8\times128$\\
        \hline
        decov0&fc2&$3\times 3 \times 64/2$&$16\times16\times64$\\
        \hline
        decov1&decov0& $3\times 3 \times 32/2$&$32\times32\times32$\\\hline
        decov2&decov1& $3\times 3 \times 16/2$&$64\times64\times16$\\\hline
        decov3&decov2& $3\times 3 \times 8/2$&$128\times128\times8$\\\hline
        decov4&decov3& $3\times 3 \times 4/2$&$256\times256\times4$\\\hline
        decov5&cov5& $3\times 3 \times 512/2$&$16\times16\times512$\\\hline
        decov6&\tabincell{c}{decov5,\\cov4,\\decov0}&$3\times 3 \times 256/2$&$32\times32\times256$\\\hline
        decov7&\tabincell{c}{decov6,\\cov3,\\decov1}&$3\times 3 \times 128/2$&$64\times64\times128$\\\hline
        decov8&\tabincell{c}{decov7,\\cov2,\\decov2}&$3\times 3 \times 64/2$&$128\times128\times64$\\\hline
        decov9&\tabincell{c}{decov8,\\cov1,\\decov3}&$3\times 3 \times 32/2$&$256\times256\times32$\\\hline
        cov6&\tabincell{c}{decov9,\\cov0,\\decov4}&$3\times 3 \times 32/1$&$256\times256\times32$\\\hline
	\end{tabular}%
\end{table}
\ifCLASSOPTIONcaptionsoff
  \newpage
\fi




\begin{thebibliography}{IEEEtran}
\bibitem{Barnich2011ViBe}
{O. Barnich and M. Van Droogenbroeck},
{"ViBe: a universal background subtraction algorithm for video sequences"},
\emph{IEEE Trans. on Image Processing},
vol.{20},
no. {6},
pp. {1709-1724},
{2011}

\bibitem{Elgammal2002Background}
{A. Elgammal, R. Duriswami, D. Harwood, and L. S. Davis},
{Background and foreground modelling using nonparametric kernel density estimation for visual surveil},
\emph{Proceedings of the IEEE},
vol. {90},
no. {7},
pp. {1151-1163},
{2002}

\bibitem{Kim2005Real}
  {K. Kim, T. H. Chalidabhongse, D. Harwood, and L. Davis},
{Real-time foreground-background segmentation using codebook model},
  \emph{Real-Time Imaging},
  vol. {11},
  no. {3},
 pp. {172-185},
{2005}

\bibitem{Zivkovic2004Improved}
{Z. Zivkovic},
{Improved adaptive gaussian mixture model for background subtraction},
 \emph{International Conference on Pattern Recognition}, {2004}


\bibitem{DBLP:journals/corr/abs-1801-02225}
{L. A. Lim and H. Y. Keles},
{Z. Liu, K. Huang, T. Tan, et al},
{Foreground segmentation using convolutional neural networks for multiscale feature encoding},
\emph{Pattern Recognition Letter},
vol. {112},
pp. {256-262},
{2018}
	
\bibitem{DBLP:journals/tits/YangLLZCL18}
{L. Yang, J. Li, Y. Luo, Y. Zhao, and H. Cheng},
{Deep background modeling using fully convolutional network},
\emph{IEEE Transaction on Intelligent Transportation Systems},
vol. {19},
no. {1},
pp. {1-9},
{2017}

\bibitem{DBLP:journals/access/ZengZ18}
{D. Zeng and M. Zhu},
{Background subtraction using multiscale fully convolutional network},
 \emph{{IEEE} Access},
vol. {6},
{16010--16021},
{2018}

\bibitem{Jobson2002A}
{D. J. Jobson, Z. Rahman, and G. A. Woodell},
{A multiscale retinex for bridging the gap between color images and the human observation of scenes},
 \emph{IEEE Transactions on Image Processing},
vol. {6},
no. {7},
pp. {965-976},
{2002}

\bibitem{Ketcham1976Real}
{D. J. Ketcham},
{Real-time image enhancement techniques},
 \emph{Osa Image Processing},
vol. {74},
no. {2},
pp. {120-125},
{1976}

\bibitem{Yan2016Nighttime}
{G. Yan, Y. Lee, and T. Q. Nguyen},
{Nighttime image enhancement applying dark channel prior to raw data from camera},
 \emph{Soc Design Conference}, {2016}

\bibitem{Cai2006Context}
{Y. Cai, K. Huang, T. Tan, and Y. Wang},
{Context enhancement of nighttime surveillance by image fusion},
 \emph{International Conference on Pattern Recognition},
{2006}

\bibitem{Liang2012Image}
{W. Liang, K. Murari, Y. Y. Zhang, Y. Chen, X. D. Li, and M. J. Li},
{Image-based fusion for video enhancement of night-time surveillance},
 \emph{Optical Engineering},
vol. {49},
no. {12},
pp. {120501-120501-3},
{2012}

\bibitem{Raskar2004Image}
{R. Raskar, A. Ilie, and J. Yu},
{Image fusion for context enhancement and video surrealism},
 \emph{International Symposium on Non-Photorealistic Animation and Rendering},
{2004}

\bibitem{goodfellow2014generative}
{I. Goodfellow, J. Pouget-Abadie, and M. Mirza et al.},
{Generative adversarial nets},
 \emph{Advances in Neural Information Processing Systems},
{2014}.

\bibitem{iizuka2017globally}
{S. Iizuka, E. Simo-Serra, and H. Ishikawa},
{Globally and locally consistent image completion},
 \emph{ACM Trans. Graph.},
vol. {36},
no. {4},
pp. {1-14},
{2017}

\bibitem{yeh2016semantic}
{R. Yeh, C. Chen, T. Y. Lim, and M. Hasegawa-Johnson et al.},
{Semantic image inpainting with perceptual and contextual losses},
 \emph{IEEE Conference on Computer Vision and Pattern Recognition},
{2016}

\bibitem{taigman2016unsupervised}
{Y. Taigman, A. Polyak, and L. Wolf},
{Unsupervised cross-domain image generation},
 \emph{International Conference on Learning Representations},
{2016}

\bibitem{ledig2017photo}
{C. Ledig, L. Theis, F. Husz′ar, J. Caballero, A. Cunningham, A. Acosta, A. P. Aitken, A. Tejani, J. Totz, and Z. Wang},
{Photo-realistic single image super-resolution using a generative adversarial network.},
 \emph{IEEE Conference on Computer Vision and Pattern Recognition},
{2017}



\bibitem{Zheng2019}
{S. Zheng, Z. Zhu, J. Cheng, Y. Guo, and Y. Zhao},
{Edge heuristic GAN for non-uniform blind deblurring},
 \emph{IEEE Signal Processing Letters},
vol. {26},
No.{10}
{1546-1550},
{2019}
\bibitem{Pixel2Pixel2017}
{P. Isola, J. Y. Zhu, T. H. Zhou, and A. A. Efros},
{Image-to-image translation with conditional adversarial networks},
 \emph{IEEE International Conference on Pattern Recognition}
{2017}



	
\bibitem{meng2019}
{Y. Meng, D. Kong, Z. Zhu, and Y. Zhao},
{From night to day: GANs based low quality image
enhancement},
\emph{Neural Processing Letters},
vol. {50},
No.{1}
{799-814},
{2019}

\bibitem{Kaewtrakulpong2002An}
{P. Kaewtrakulpong and R. Bowden},
{An improved adaptive background mixture model for real-time tracking with shadow detection},
 \emph{Video-Based Surveillance Systems, Springer, Berlin},
pp. {135-144},
{2002}

\bibitem{Stauffer1999Adaptive}
{P. Kaewtrakulpong and R. Bowden},
{Adaptive background mixture models for real-time tracking},
 \emph{IEEE Conference Computer Vision and Pattern Recognition},
{1999}

\bibitem{Elgammal2000Non}
{A. M. Elgammal, D. Harwood, and L. S. Davis},
\emph{Non-parametric model for background subtraction},
{European Conference on Computer Vision},
{2000}

\bibitem{heikkila2006texture}
{M. Heikkila and M. Pietikainen},
{A texture-based method for modeling the background and detecting moving objects},
\emph{IEEE Transactions on Pattern Analysis and Machine Intelligence},
vol. {28},
no. {4},
pp. {657--662},
{2006}

\bibitem{DBLP:conf/cvpr/LiaoZKPL10}
{S. Liao, G. Zhao, and V. Kellokumpu et al.},
{Modeling pixel process with scale invariant local patterns for background subtraction in complex scenes},
\emph{IEEE Conference on Computer Vision and Pattern Recognition},
{2010}

\bibitem{arjovsky2017wasserstein}
{M. Arjovsky, S. Chintala, and L. Bottou},
{Wasserstein generative adversarial networks},
\emph{International Conference on Machine Learning}, {2017}
\bibitem{he2016deep}
{K. He, X. Zhang, S. Ren, and J. Sun},
{Deep residual learning for image recognition},
 \emph{IEEE Conference on Computer Vision and Pattern Recognition},
{2016}
\bibitem{gatys2016image}
  {L. A. Gatys, A. S. Ecker, and M. Bethge},
    {Image style transfer using convolutional neural networks},
\emph{IEEE Conference on Computer Vision and Pattern Recognition}, {2016}

\bibitem{st2014improving}
  {P.L. St-Charles and G. A. Bilodeau},
{Improving background subtraction using local binary similarity patterns},
  \emph{IEEE Winter Conference on Applications of Computer Vision (WACV)},
   {2014}

\bibitem{maddalena2010fuzzy}
  {L. Maddalena and A. Petrosino},
{A fuzzy spatial coherence-based approach to background/foreground separation for moving object detection},
  \emph{Neural Computing and Applications},
 vol. {19},
  no. {2},
  pp. {179--186},
  {2010}

\bibitem{maddalena2008self}
  {L. Maddalena, A. Petrosino, et al.},
{A self-organizing approach to background subtraction for visual surveillance applications},
  \emph{IEEE Trans. on Image Processing},
 vol. {17},
 no. {7},
pp. {1168-1177},
{2008}

\bibitem{st2014flexible}
  P.-L. St-Charles, G.-A. Bilodeau, and R. Bergevin,
{Flexible background subtraction with self-balanced local sensitivity},
  \emph{IEEE Conference on Computer Vision and Pattern Recognition Workshops},
  {2014}

\bibitem{bloisi2012independent}
  {D. Bloisi and L. Iocchi},
{Independent multimodal background subtraction.},
  \emph{Computational Modelling of Objects Represented in Images Fundamentals Methods and Applications \uppercase\expandafter{\romannumeral3}},
  pp. {39--44},
  {2012}

\bibitem{godbehere2012visual}
  {A. B. Godbehere, A. Matsukawa, and K. Goldberg},
{Visual tracking of human visitors under variable-lighting conditions for a responsive audio art installation},
 \emph {American Control Conference},
    {2012}


\bibitem{zhao2012fuzzy}
{Z. Zhao, T. Bouwmans, X. Zhang, and Y. Fang},
{A fuzzy background modeling approach for motion detection in dynamic backgrounds},
\emph{Multimedia and Signal Processing},
{2012}

\bibitem{rosenfeld1970edge}
{Z. Zhu, H. Lu, and Y. Zhao},
{Scale multiplication in odd Gabor transform domain for edge detection},
\emph{Journal of Visual Communication and Image Representation},
vol. {18},
no. {1},
pp. {68-80},
{2007}

\bibitem{rosenfeld1971edge}
{L. Zhang and P. Bao},
{Edge detection by scale multiplication in wavelet domain},
\emph{Pattern Recognition Letters},
vol. {23},
no. {14},
pp. {1771-1784},
{2002},

\bibitem{liu2012foreground}
{Z. Liu, K. Huang, T. Tan, et al},
{Foreground object detection using top-down information based on EM framework},
\emph{IEEE Transaction on Image Processing},
vol. {21},
no. {9},
pp. {4204-4217},
{2012}



\bibitem{bgslibrary}
{A. Sobral},
{BGSLibrary: An OpenCV C++ background subtraction Library},
\emph{IX Workshop de Vison Computacional (WVC'2013)},
{Rio de Janeiro, Brazil},
{2013}

\bibitem{Noh2012A}
{S. J. Noh and M. Jeon},
{A new framework for background subtraction using multiple cues},
\emph{Asian Conference on Computer Vision},
{2012}

\bibitem{jia2017improving}
{K. Jia, D. Tao, S. Gao, and X. Xu},
{Improving training of deep neural networks via singular value bounding},
\emph{IEEE Conference on Computer Vision and Pattern Recognition},
{2017}


\bibitem{yu2017compressing}
{X. Yu, T. Liu, X. Wang, and D. Tao},
{On compressing deep models by low rank and sparse decomposition},
\emph{IEEE Conference on Computer Vision and Pattern Recognition},
{2017}

\bibitem{DBLP:Guo2016}
{X. Guo, L. Yu, and H. Ling},
{LIME: Low-light image enhancement via illumination map estimation},
\emph{IEEE Transactions on image processing},
vol. {26},
no. {2},
pp. {982-993},
{2016}

\bibitem{DBLP:Lore2017}
{K. Lore, A. Adedotun, and S. Soumik},
{LLNet: A deep autoencoder approach to natural low-light image enhancement},
\emph{Pattern Recognition},
vol. {61},
pp. {650-662},
{2017}

\bibitem{DBLP:Li2018}
{M. Li, J. Liu, W. Yang, X. Sun, and Z. Guo},
{Structure-revealing low-light image enhancement via robust retinex model},
\emph{IEEE Trans. on Image Processing},
vol. {27},
no. {6},
pp. {2828-2841},
{2018}
\end{thebibliography}
%

%
\begin{IEEEbiography}[{\includegraphics[width=1in,height=1.25in,clip,keepaspectratio]{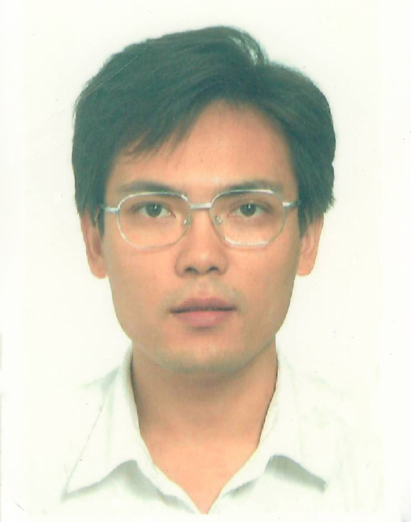}}]{Zhenfeng Zhu}
	received the M.E. degree from Harbin Institute of Technology, Harbin, China, in 2001, and the Ph.D. degree from National Laboratory of Pattern Recognition (NLPR), Institute of Automation, CAS, Beijing, China, in 2005, respectively. He was a Visiting Scholar with the Department of Computer Science and Engineering, Arizona State University, AZ, USA, in 2010. He is currently a Professor with School of Computer and Information Technology, Beijing Jiaotong University, Beijing, China.

He has authored or co-authored over 100 papers in journals and conferences including IEEE T-KDE, T-NNLS, T-CSVT, T-MM, T-CYB, CVPR, AAAI, IJCAI, and ACM Multimedia etc. His group has won the Honorable Award (Rank 1st in stage two) in KDD Cup2016 competition and Rank3 in CIKM Cup2016 competition, respectively. His current research interests include image and video understanding, computer vision, machine learning, and recommendation system.
\end{IEEEbiography}

\vspace{-10 mm}
\begin{IEEEbiography}[{\includegraphics[width=1in,height=1.25in,clip,keepaspectratio]{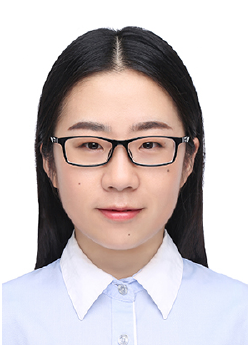}}]{Yingying Meng}
received a B.S. degree in Computer Science and Technology from  Shandong University of Science and Technology, Qingdao, China, in 2016. She received the M.E. degree in Signal and Information Processing from Institute of Information Science, Beijing Jiaotong University, Beijing, China, in 2019.  She is currently a software engineer in China CITIC Bank.
\end{IEEEbiography}

\vspace{-15 mm}
\begin{IEEEbiography}[{\includegraphics[width=1in,height=1.25in,clip,keepaspectratio]{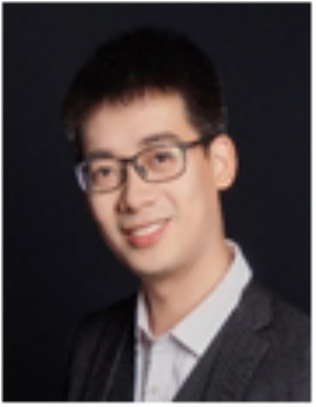}}]
	{Deqiang Kong} received the M.E. degree in signal and information processing from Institute of Information Science, Beijing Jiaotong University, Beijing, China, in 2018. He is currently an applied scientist with Microsoft, Beijing, China. His research interests include video understanding, natural language processing and recommendation system.

\end{IEEEbiography}

\vspace{-15 mm}
\begin{IEEEbiography}[{\includegraphics[width=1in,height=1.25in,clip,keepaspectratio]{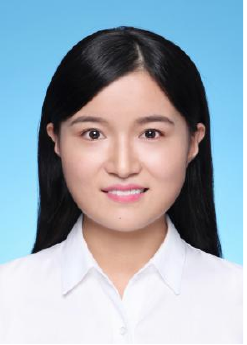}}]{Xingxing Zhang}
received the B.S. degree in communication engineering from Henan Normal University, Xinxiang, China, in 2015. Currently, she is pursuing the Ph.D. degree in the Institute of Information Science, Beijing Jiaotong University, Beijing, China.

Her research interests include data analysis, image and video understanding, computer vision, and machine learning.
\end{IEEEbiography}

\vspace{-15 mm}
\begin{IEEEbiography}[{\includegraphics[width=1in,height=1.25in,clip,keepaspectratio]{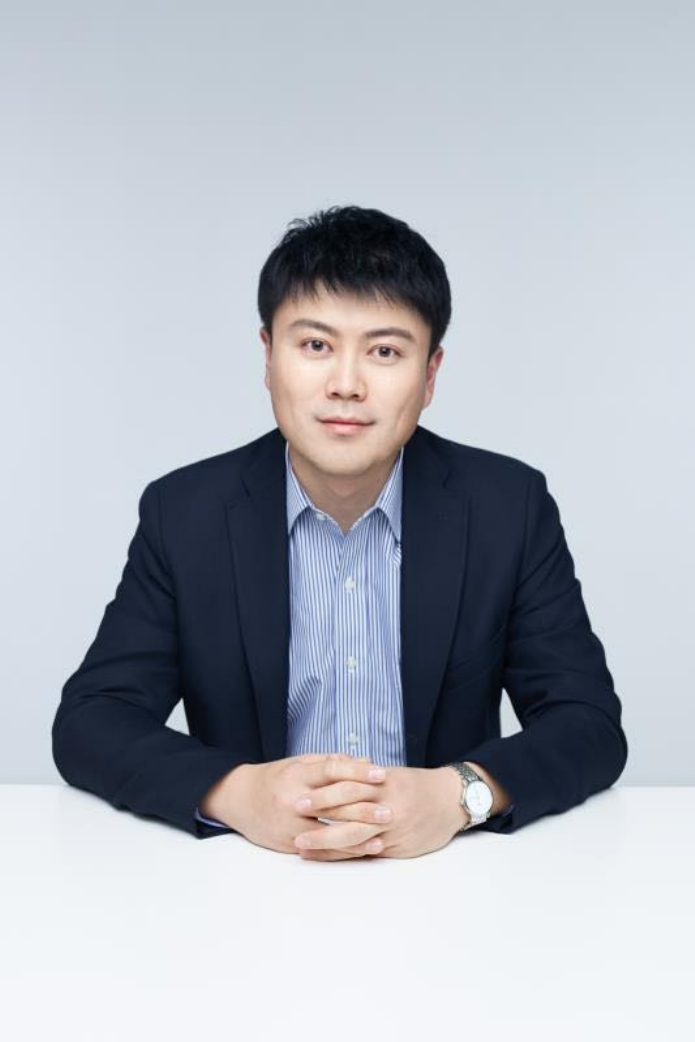}}]{Yandong Guo}
 earned his Ph.D. in ECE from Purdue University at West Lafayette, under the supervision of Prof. Bouman and Prof. Allebach in 2013. Before that, he received his B.S. and M.S. degree in ECE from Beijing University of Posts and Telecommunications, China, in 2005 and 2008, receptively.
He is the chief scientist of intelligent perception at OPPO Research Institute. He also holds an adjunct professor position in Beijing University of Posts and Telecommunications, and University of Electronic Science and Technology of China. Before he joined OPPO in 2020, he was the chief scientist at XPeng Motors (China) and a researcher at Microsoft Research, Redmond WA in the United States. His professional interests lie in the broad area of computer vision, imaging system, human behavior understanding and biometric, and autonomous driving.

\end{IEEEbiography}

\vspace{-15 mm}
\begin{IEEEbiography}[{\includegraphics[width=1in,height=1.25in,clip,keepaspectratio]{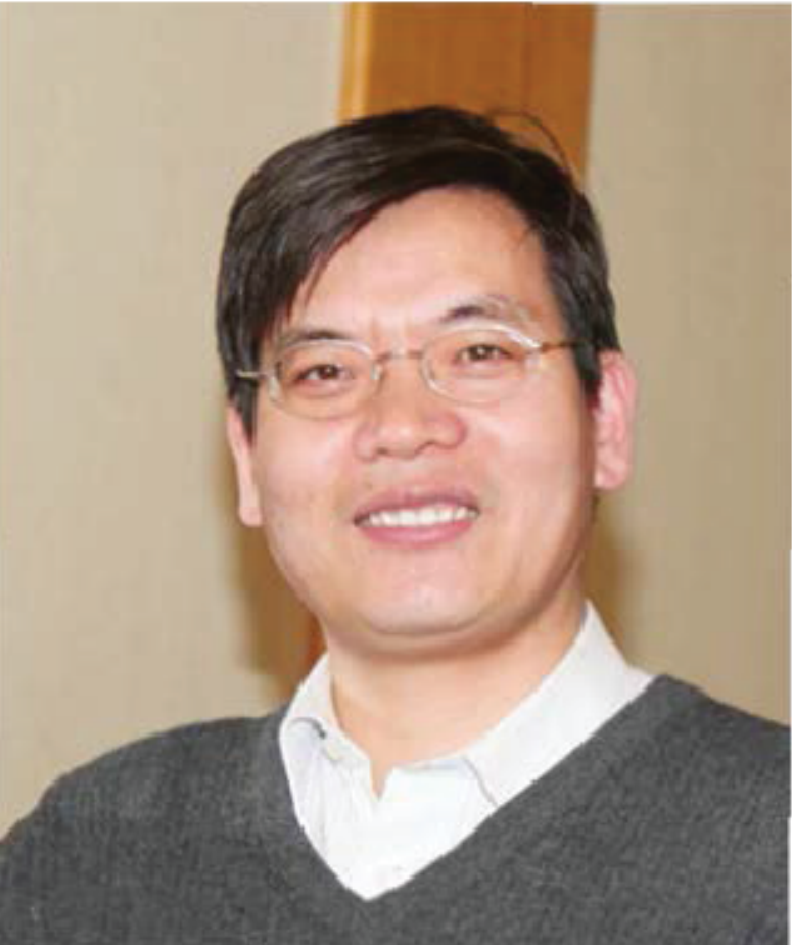}}]{Yao Zhao}
received the B.S. degree from Fuzhou University, Fuzhou, China, in 1989, the M.E. degree from Southeast University, Nanjing, China, in 1992, and the Ph.D. degree from the Institute of Information Science, Beijing Jiaotong University (BJTU), Beijing, China, in 1996.

From 2001 to 2002, he was a Senior Research Fellow with the Information and Communication Theory Group, Delft University of Technology, Delft, The Netherlands. Since 2001, he has been a Professor with BJTU. He is currently the Director of the Institute of Information Science, BJTU. His current research interests include image/video coding, digital watermarking and forensics, and video analysis and understanding.

He serves on the editorial boards of several international journals, including as an Associate Editor of the IEEE Trans. on Cybernetics, Associate Editor of the IEEE Signal Processing Letters, the Area Editor of Signal Processing: Image Communication, and an Associate Editor of Circuits, System, and Signal Processing. He was named a Distinguished Young Scholar by the National Science Foundation of China in 2010, and was elected as a Chang Jiang Scholar of the Ministry of Education of China in 2013.
\end{IEEEbiography}

%
%
%




\end{document}